\PassOptionsToPackage{dvipsnames,table}{xcolor}
\documentclass{article}

\PassOptionsToPackage{nopatch=eqnum}{microtype}
\PassOptionsToPackage{hyphens}{url}
\usepackage{aibl_arxiv}
\usepackage{amsmath,amsfonts,bm}

\usepackage{graphicx}
\usepackage{xcolor}
\usepackage{booktabs,tabularx,array,multirow}
\usepackage{amsmath,amssymb}
\usepackage{url}
\usepackage{placeins}
\usepackage{titletoc}
\AddToHook{env/figure/begin}{%
  \setlength{\abovecaptionskip}{3.5pt}%
  \setlength{\belowcaptionskip}{0pt}%
}
\AddToHook{env/table/begin}{%
  \setlength{\abovecaptionskip}{0pt}%
  \setlength{\belowcaptionskip}{3pt}%
}
\AddToHook{env/table*/begin}{%
  \setlength{\abovecaptionskip}{0pt}%
  \setlength{\belowcaptionskip}{3pt}%
}
\graphicspath{{figures/}}
\makeatletter\@secpenalty=0
\renewcommand\section{\@startsection{section}{1}{\z@}%
  {-2.0ex \@plus -1ex \@minus -.2ex}{1.0ex \@plus .2ex}%
  {\sffamily\bfseries\large\leavevmode\color{aiblteal}}}
\renewcommand\subsection{\@startsection{subsection}{2}{\z@}%
  {-1.6ex \@plus -1ex \@minus -.2ex}{0.8ex \@plus .2ex}%
  {\sffamily\bfseries\normalsize\leavevmode\color{aiblteal}}}
\makeatother
\newcommand{\Mzero}{M_0}
\newcommand{\aiblresourcelink}[3]{{\sffamily\bfseries\small #1:} {\small\href{#2}{\nolinkurl{#3}}}\par}
\newcommand{\runinsubsection}[2]{\par\refstepcounter{subsection}\noindent\label{#2}\textbf{\thesubsection\enspace #1.}\hspace{0.5em}\ignorespaces}

\providecommand{\venue}[1]{%
  \hspace{0.35em}{\tiny\textcolor{black!55}{#1}}%
}
\renewcommand{\eqref}[1]{\textup{(\ref{#1})}}

\begin{document}

\aiblmaketitle
  {Edited Endpoints as Partial Goals for\\Image-to-Video Generation}
  {Sihyun Kim$^{1}$, \ Sanghyun Jo$^{1,2,\dagger}$, \ Kyungsu Kim$^{1,\dagger}$}
  {$^{1}$Seoul National University \quad $^{2}$OGQ\\[0.15em]
   {\small $^{\dagger}$Corresponding authors: \texttt{shjo.april@gmail.com}, \texttt{kyskim@snu.ac.kr}}}
  {Edited endpoint images make the target of an image-to-video transition explicit and can improve transition completion, but also preserve context in the source image's layout. Full-frame endpoint conditioning turns this preserved context into requirements on the video's final state, potentially inducing temporal artifacts. We measure discontinuity with Hard-Cut (HC), late background reversal with Background Motion Reversal (BMR), and motion suppression with Dynamic Degree (DD). These artifacts appear across five video models spanning training-free and learned endpoint conditioning, in model-specific forms. We study Partial, a selective conditioning scheme that constrains only an edit-relevant region estimated from image-editor attention and leaves the remaining terminal content to the video model. Across three backbones on TC-Bench-I2V, Partial reduces temporal artifacts relative to full-frame conditioning, with no significant difference in Transition Completion Ratio (TCR); on \mbox{Wan2.2-Turbo}, uniformly weakening conditioning does not reproduce these reductions. A donor-context intervention keeps full-frame conditioning and replaces only the context outside the edit-relevant region with the final frame of a Partial video. It lowers HC and BMR on HunyuanVideo-1.5 and lowers BMR while restoring DD on \mbox{Wan2.2-Turbo}, with no significant difference in TCR, so constraint extent alone does not account for these artifacts. Together, these findings suggest that edited endpoints should serve as partial future goals, although edit relevance does not guarantee future validity. Code and data will be released at the link below.}
  {\raggedright\aiblresourcelink{Project}{https://wannagetacat.github.io/Edited-Endpoints-I2V/}{wannagetacat.github.io/Edited-Endpoints-I2V}
   \aiblresourcelink{Github}{https://github.com/wannagetacat/Edited-Endpoints-I2V}{github.com/wannagetacat/Edited-Endpoints-I2V}
   \aiblresourcelink{Lab}{https://aibl.snu.ac.kr}{aibl.snu.ac.kr}}

\section{Introduction}
\label{sec:intro}

Given a source image, ``change the bus to purple'' specifies a target appearance but not a trajectory. A video may miss or undo the change before its last frame \citep{tcbench}. Editing the source image makes the target explicit, as edited keyframes already do in video editing and interpolation \citep{anyv2v,i2vedit,genprop,maskint,mfm}.

An edited endpoint mixes two kinds of information: the purple bus specifies the target appearance, while the unchanged road, buildings, and sign preserve the editor's input in the source image's layout. When an edited frame shares its source frame's time, as in edit-then-propagate video editing, this alignment makes it a reliable anchor. As an endpoint, the same context is imposed at a later time: full-frame terminal conditioning requires the road, buildings, and sign to return to their source-time positions at the video's end. We call this promotion of preserved context to spatial future requirements \emph{spatial over-specification}.

Consistent with this concern, full-frame endpoint conditioning can improve completion while suppressing motion or increasing discontinuity and background motion reversal, under both training-free projection and learned endpoint conditioning (Tables~\ref{tab:main} and~\ref{tab:learned}). We study \emph{whether terminal content affects temporal artifacts beyond constraint extent}.

\begin{figure}[t]
\centering
\includegraphics[width=\linewidth,trim=0 3pt 0 3pt,clip]{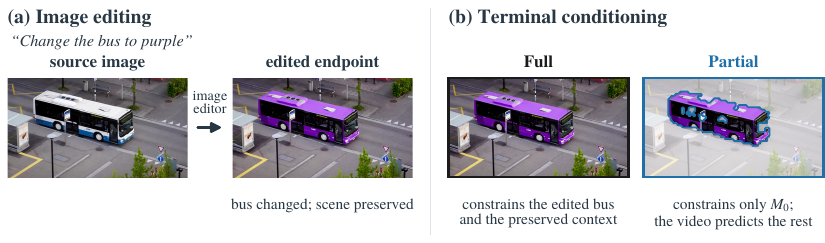}
\caption{\textbf{Edited endpoints preserve context that need not specify the video's future.} Full-frame endpoint conditioning constrains both requested appearance and preserved context. Partial constrains only estimated edit-relevant support $M_0$ from image-editor attention (blue outline); washed-out positions retain the video prediction. Source frame: DAVIS \emph{bus} \citep{davis}, via FiVE-Bench \citep{five}.}
\label{fig:support_overview}
\end{figure}

Using established masked conditioning \citep{repaint,blendedlatent}, we test content at fixed full-frame support and vary support area, placement, and global mixing weight. Partial constrains the terminal frame to the edited endpoint only within fixed edit-relevant support estimated from image-editor attention, retaining the video's prediction elsewhere (Fig.~\ref{fig:support_overview}).

Full-frame endpoint conditioning and Partial differ in both the imposed terminal content and the constraint extent. To test content at fixed extent, we run a donor-context intervention: inside the edit-relevant support $M_0$, we keep the original edited endpoint; outside $M_0$, we substitute the final frame of a paired Partial rollout, the donor, and constrain the entire terminal frame with the same full-frame projection.

\begin{samepage}
Across 103 TC-Bench-I2V prompts with five paired seeds per condition, we report three findings:\par\nopagebreak

\textbf{1. Terminal content matters beyond constraint extent.} With full-frame support unchanged, the donor-context intervention lowers Hard-Cut (HC) and Background Motion Reversal (BMR) scores on HunyuanVideo-1.5, and lowers BMR while restoring Dynamic Degree (DD) on Wan2.2-Turbo (Fig.~\ref{fig:context}).
\end{samepage}

\textbf{2. Support area and placement yield distinct completion--temporal trade-offs.} On Wan2.2-Turbo, edit-relevant placement favors completion at matched area; shrinking support lowers HC, whereas enlarging it raises BMR. Tested global softening does not reproduce Partial's artifact reductions.

\textbf{3. Partial reduces temporal artifacts at comparable completion.} Across three backbones, Partial mitigates temporal artifacts relative to full-frame endpoint conditioning at comparable Transition Completion Ratio (TCR), and improves completion over Endpoint-free on Wan2.2-Turbo and SANA-Video. Restricting gradient-based Frame Guidance \citep{frameguidance} to the same support likewise keeps its completion gain and removes its motion suppression. Partial's residual failures show that edit relevance need not imply future-valid position or visibility.

\section{Related Work}
\label{sec:related}

\par\noindent\textbf{Frame guidance and masked conditioning.}
Frame Guidance optimizes video latents using selected decoded frames, including edited-frame inputs and binary-masked $L_2$ guidance \citep{frameguidance}. Our projection follows RePaint's mixing of generated and noise-matched known pixels and Blended Latent Diffusion's mixing of generated and source-image latents \citep{repaint,blendedlatent}. We use this established mixing principle to analyze terminal constraints, rather than introduce a new guidance operator. Projection lets us vary terminal content, constraint extent, and placement within one fixed sampler. With its frame loss restricted to the same edit-relevant support, Frame Guidance also keeps its completion gain and recovers the motion that full-frame guidance suppresses (Section~\ref{sec:fg}).

\par\noindent\textbf{Edit localization and regional control.}
Prompt-to-Prompt connects prompt words with image layout through cross-attention \citep{prompttoprompt}. DiffEdit derives masks from contrasts between prompt-conditioned noise predictions, whereas InstDiffEdit derives attention masks during editing \citep{diffedit,instdiffedit}. MasaCtrl and MAG-Edit use masks for attention control and local latent optimization, respectively \citep{masactrl,magedit}. EditCtrl combines local masked and global temporal context for video editing \citep{editctrl}. For video generation from a single image, we use image-editor attention to estimate edit-relevant support on the terminal frame.

\par\noindent\textbf{Edited keyframes, target states, and inbetweening.}
Prior work uses edited or specified frames as appearance anchors to propagate or as states to connect, typically imposing the whole frame at its time index. We instead impose an edited frame at a different time from its source, with no trajectory relating the two, and ask which parts of its spatial content remain valid future requirements. Edit-then-propagate video editing edits the first frame with an image editor and carries it along a source video that supplies motion \citep{anyv2v,i2vedit,genprop}. TC-Bench's SDXL+SEINE baseline and MaskINT edit keyframes and interpolate between them \citep{tcbench,sdxl,seine,maskint}. Inbetweening methods connect specified start and end states through fine-tuned or training-free bidirectional generation \citep{generative_inbetweening,trf,vibidsampler,vtg}. Wan2.1-FLF2V, Many-for-Many, and OmniWeaving instead learn first--last-frame or flexible frame conditioning \citep{wan,mfm,omniweaving}. VACE, Many-for-Many, and VideoCanvas support spatially masked video conditioning \citep{vace,mfm,videocanvas}, offering learned approaches to partial goals.

\par\noindent\textbf{Temporal artifacts of conditioning.}
Conditional image leakage links limited I2V motion to source-image over-reliance at large diffusion timesteps \citep{imageleakage}. InfinityStory's supplementary ablations report restricted natural motion from fixing each shot's keyframe and rushed transitions toward prescribed positions \citep{infinitystory}. DCARL addresses keyframe duplication and boundary discontinuity with noisy conditioning and boundary consistency \citep{dcarl}. TIC-FT diagnoses abrupt content and noise-level transitions under direct concatenation and uses graded-noise buffer frames \citep{ticft}. Building on these observations, our donor-context intervention tests whether replacing editor-preserved terminal context changes temporal artifacts under unchanged full-frame support and projection (Section~\ref{sec:context}).

\section{Problem Formulation and Method}
\label{sec:problem}

\subsection{Problem Setting}

Given a source image $I_0$ and a transition prompt, an image editor produces an endpoint $I_E$ from an edit prompt that embeds the transition prompt and asks for the frame after the transition completes (Appendix~\ref{app:implementation}). A frozen video model generates $V=(I_0,\ldots,I_T)$ from the source and the same transition prompt under three conditions: Endpoint-free, full-frame endpoint conditioning (Full), and partial endpoint conditioning (Partial). Within each controlled comparison, conditions share the source and prompt; endpoint-conditioned conditions share $I_E$ except in Section~\ref{sec:context}.

\subsection{Edited Endpoints and Partial Goals}

An idealized decomposition makes the mismatch explicit:
\begin{equation}
I_E \approx M^*\odot I_T^{\mathrm{edit}}+(1-M^*)\odot I_0,
\label{eq:decomposition}
\end{equation}
where $M^*$ is an idealized edit-relevance mask and $I_T^{\mathrm{edit}}$ denotes the requested target appearance. This decomposition represents an editor that changes requested content while preserving context in the source layout. The second term adds no appearance beyond $I_0$, which already conditions the video; as a terminal requirement, it only fixes where source content must be at time $T$. Only the first term carries information that $I_0$ lacks. Full treats both terms as future requirements. Yet preserved context may change position or visibility as the scene evolves, so forcing it back can impose constraints beyond the edit prompt: spatial over-specification. Our FLUX.2 endpoints follow this pattern: outside the estimated support $\Mzero$ (Section~\ref{sec:support}), they differ from $I_0$ by a mean absolute RGB difference of 0.056, versus 0.232 inside (Appendix~\ref{app:layout}).

\par\noindent\textbf{Edit relevance and future validity.}
\emph{Edit relevance} associates content with the requested change; \emph{future validity} concerns its appropriateness as a future constraint. Partial uses the former to select support, without estimating the latter. Neither $M^*$ nor its estimate below is a future-validity mask.

\subsection{Edit-Relevant Support Estimation}
\label{sec:support}

We estimate binary support $\Mzero$ from the editor's text--image attention to object and edit phrases preselected from each prompt and fixed across conditions. We normalize and average their attention maps, then apply a two-component Gaussian mixture with an Ising spatial prior (GMM--Ising; Appendix~\ref{app:implementation}; Fig.~\ref{fig:support_examples}).

The object and edit phrases connect the requested change to spatial regions through the editor's attention, so $\Mzero$ approximates $M^*$ in Eq.~\eqref{eq:decomposition} without a separate segmentation model.

\subsection{Full, Soft, and Partial Conditioning}
\label{sec:method}

Partial imposes the endpoint only on $\Mzero$, keeping the content that $I_0$ lacks, and retains the video prediction elsewhere because future context is unknown.

\par\noindent\textbf{Weighted terminal projection.}
Let $z_E=E(I_E)$ encode the endpoint, $u$ index spatial latent positions, and $\hat z^{(k)}_{T}$ denote the clean prediction of the terminal latent frame at sampling step $k$. All conditions share one projection with a spatial weight $w$:
\begin{equation}
\hat z^{(k)}_{T}(u) \leftarrow
w(u)\,z_E(u)+[1-w(u)]\,\hat z^{(k)}_{T}(u),
\label{eq:partial}
\end{equation}
with $w\equiv1$ for Full, $w\equiv\alpha\in\{0.75,0.5,0.25\}$ for Soft, and $w=\Mzero$ for Partial, after transformation to each latent grid. Soft weakens every terminal position uniformly; $\alpha$ is a mixing weight, not a calibrated measure of overall constraint strength. Support stays fixed in terminal spatial coordinates throughout sampling and does not track motion.

\par\noindent\textbf{Backbone implementations.}
Wan2.2 applies the same $w$ to the terminal model input, with token timesteps scaled by $1-w$. SANA Partial converts the projected clean prediction back to its solver's flow, while SANA Full clamps the terminal state. Hunyuan uses a global timestep, so it projects toward the endpoint noised to the current level after each scheduler update. Appendix~\ref{app:implementation} specifies encoding, mask transforms, and noise handling. All controlled comparisons are paired within a backbone, so these implementation differences do not enter them. Learned endpoint-conditioning models are additional observational baselines, not implementations of this projection.

\section{Main Experiments}
\label{sec:experiments}

\runinsubsection{Experimental Setup and Metrics}{sec:setup}
We evaluate the 103 of TC-Bench-I2V's 120 prompts whose source videos remain available, with five paired seeds per condition (515 videos) \citep{tcbench}, using FLUX.2 endpoints; prompt IDs follow TC-Bench. The primary backbones are Wan2.2-Turbo (Wan2.2), SANA-Video 2B \citep[SANA;][]{sana}, and HunyuanVideo-1.5 \citep[Hunyuan;][]{hunyuan}. Wan2.2-Turbo is quanhaol's community-distilled \mbox{Wan2.2-TI2V-5B-Turbo} checkpoint.\footnote{Official releases: \href{https://huggingface.co/black-forest-labs/FLUX.2-klein-base-9B}{FLUX.2-klein-base-9B} and \href{https://github.com/quanhaol/Wan2.2-TI2V-5B-Turbo}{Wan2.2-TI2V-5B-Turbo}.} Appendix~\ref{app:implementation} gives sampler configurations.

\par\noindent\textbf{Image-editing prior and comparison logic.}
FLUX.2-klein-base-9B edits the source image into a plausible final frame after the requested transition has completed (Appendix~\ref{app:editor}). We generate one endpoint per prompt and reuse it for Full and Partial. Endpoint-free receives the source image and transition prompt; Full additionally constrains the entire terminal frame to the edited endpoint, whereas Partial constrains only $M_0$. Thus Endpoint-free versus Full tests the completion benefit and temporal artifacts of adding an explicit image-editing goal, while Full versus Partial tests selective use of that same goal. Partial requires no video-model fine-tuning.

\par\noindent\textbf{Related baselines.}
Table~\ref{tab:main} also summarizes selected comparisons from Tables~\ref{tab:tcbench_i2v} and~\ref{tab:learned}. Within the Wan2.2 family, we also include the non-distilled Wan2.2 Base and Wan2.2 with Qwen-Image-Edit endpoints \citep{wan,qwenimage}. OmniWeaving, built on HunyuanVideo-1.5, and MfM provide learned endpoint-conditioning baselines \citep{omniweaving,mfm}. The three controlled blocks compare conditions within each backbone; the additional methods use their native configurations, with edit-type breakdowns and timing in Table~\ref{tab:tcbench_i2v}.

\par\noindent\textbf{Completion and temporal diagnostics.}
We report Transition Completion Ratio (TCR) and TC-Score with a Qwen2.5-VL-32B-Instruct judge \citep{qwen25vl}. TCR conjoins transition and object-consistency assertions, excluding ``other objects'' as in TC-Bench \citep{tcbench}. TC-Score excludes transition-object consistency from its VLM term and adds adjacent-frame CLIP consistency \citep{clip}, which also varies with motion, so we use TCR as the primary completion measure.

Temporal diagnostics use a common 8 FPS grid. Dynamic Degree (DD) detects substantial movement using VBench's RAFT-based decision \citep{vbench,raft}. Discontinuities and camera drift can raise flow scores \citep{salt,steadyforcing}, so DD does not assess motion appropriateness. Hard-Cut (HC) measures the largest drop of the adjacent-frame similarity $s_t$ below its median, where $\phi$ is the CLIP image encoder and $I_t$ the $t$-th frame on this grid:
\begin{equation}
\mathrm{HC}=\operatorname{median}_t s_t-\min_t s_t,
\qquad
s_t=\cos\big(\phi(I_t),\phi(I_{t+1})\big).
\label{eq:hc}
\end{equation}

BMR measures the fraction of late along-axis motion mass opposing an early reference direction in the region outside the original estimated support $\Mzero$, which approximates the background and stays fixed across conditions, including shifted/random conditioning supports. Let $F_t(x)$ denote WAFT flow \citep{waft} and $L$ the number of flow pairs; the median is taken componentwise. The magnitude-weighted vector $b_t$, reference axis $\hat b$, and reversal score are
\begin{align}
b_t&=\frac{\sum_{x\notin\Mzero}\|F_t(x)\|F_t(x)}{\sum_{x\notin\Mzero}\|F_t(x)\|},
&\hat b&=\frac{b_{\rm ref}}{\|b_{\rm ref}\|},\quad b_{\rm ref}=\operatorname{median}_{t<L/2} b_t,\nonumber\\
\mathrm{BMR}&=\frac{\sum_{t\ge L/2}\max(0,-\langle b_t,\hat b\rangle)}{\sum_{t\ge L/2}|\langle b_t,\hat b\rangle|}.
\label{eq:bmr}
\end{align}
BMR lies in $[0,1]$ and depends on the estimated mask and flow. Section~\ref{sec:validation} gives track-based checks.

\par\noindent\textbf{Uncertainty.}
We report 95\% confidence intervals (CIs) from 100,000 paired prompt-bootstrap draws over 103 prompts (Appendix~\ref{app:statistics}). A paired difference is significant when its pointwise CI excludes zero; otherwise we call the conditions comparable, which does not imply equivalence.

\begin{table}[!t]
\centering
\setlength{\tabcolsep}{3pt}
\caption{\textbf{Image-editing goals: completion and temporal trade-offs.} TC-Bench-I2V, 103 prompts $\times$ five paired seeds. Endpoint-free uses the source image and prompt only. Full and Partial share the same FLUX.2-klein-base-9B edited endpoint; their constraints cover the entire terminal frame and $M_0$, respectively. Related baselines use the endpoint prior named in each row.}
\label{tab:main}
\fitbox{%
\begin{tabular}{@{}lrrrrrr@{}}
\toprule
 & \multicolumn{3}{c}{Transition completion} & \multicolumn{3}{c}{Temporal diagnostics} \\
\cmidrule(lr){2-4} \cmidrule(l){5-7}
Condition / endpoint prior & TCR $\uparrow$ & $\Delta$TCR (pp) $\uparrow$ & TC-Score $\uparrow$ & HC $\downarrow$ & DD $\uparrow$ & BMR $\downarrow$ \\
\midrule
\multicolumn{7}{@{}l}{\textbf{Wan2.2-Turbo}} \\
Endpoint-free & 33.79 & --\hphantom{$^{*}$} & 0.7841 & \textbf{4.57} & \textbf{79.4} & \textbf{0.360} \\
Full (FLUX.2) & \underline{41.94} & $+8.16^{*}$ & \textbf{0.8352} & 6.47 & 66.8 & 0.599 \\
Partial (FLUX.2) & \textbf{42.91} & $+9.13^{*}$ & \underline{0.8187} & \underline{5.76} & \underline{76.7} & \underline{0.387} \\
\rowcolor{ForestGreen!12}$\Delta$ (Partial $-$ Full) & $+0.97\hphantom{{}^{*}}$ & --\hphantom{$^{*}$} & $-0.0164^{*}$ & $-0.72^{*}$ & $+9.90^{*}$ & $-0.213^{*}$ \\
\midrule
\multicolumn{7}{@{}l}{\textbf{SANA-Video}} \\
Endpoint-free & 23.50 & --\hphantom{$^{*}$} & 0.7530 & \textbf{8.07} & \textbf{42.7} & \textbf{0.122} \\
Full (FLUX.2) & \textbf{35.73} & $+12.23^{*}$ & \textbf{0.7944} & 9.77 & 32.8 & 0.420 \\
Partial (FLUX.2) & \underline{35.34} & $+11.84^{*}$ & \underline{0.7935} & \underline{8.80} & \underline{42.3} & \underline{0.135} \\
\rowcolor{ForestGreen!12}$\Delta$ (Partial $-$ Full) & $-0.39\hphantom{{}^{*}}$ & --\hphantom{$^{*}$} & $-0.0008\hphantom{{}^{*}}$ & $-0.97\hphantom{{}^{*}}$ & $+9.51^{*}$ & $-0.285^{*}$ \\
\midrule
\multicolumn{7}{@{}l}{\textbf{HunyuanVideo-1.5}} \\
Endpoint-free & 40.97 & --\hphantom{$^{*}$} & \underline{0.8718} & \textbf{3.47} & \underline{51.5} & \textbf{0.305} \\
Full (FLUX.2) & \textbf{46.60} & $+5.63\hphantom{{}^{*}}$ & 0.8588 & 12.96 & \textbf{52.6} & 0.585 \\
Partial (FLUX.2) & \underline{44.47} & $+3.50\hphantom{{}^{*}}$ & \textbf{0.8804} & \underline{5.69} & \underline{51.5} & \underline{0.333} \\
\rowcolor{ForestGreen!12}$\Delta$ (Partial $-$ Full) & $-2.14\hphantom{{}^{*}}$ & --\hphantom{$^{*}$} & $+0.0216^{*}$ & $-7.27^{*}$ & $-1.17\hphantom{{}^{*}}$ & $-0.252^{*}$ \\
\midrule
\multicolumn{7}{@{}l}{\textbf{Related I2V and endpoint baselines} (Tables~\ref{tab:tcbench_i2v} and~\ref{tab:learned})} \\
Wan2.2 Base (FLUX.2) & 45.24 & $+14.17^{*}$ & 0.8475 & 7.18 & 57.9 & 0.497 \\
Wan2.2-Turbo (Qwen-Image-Edit) & 35.15 & $+1.36\hphantom{{}^{*}}$ & 0.8001 & 7.11 & 64.1 & 0.617 \\
OmniWeaving (FLUX.2) & 43.30 & $+3.88\hphantom{{}^{*}}$ & 0.8647 & 5.65 & 36.9 & 0.487 \\
MfM (FLUX.2) & 43.88 & $+11.26^{*}$ & 0.8869 & 4.13 & 28.2 & 	0.387 \\
\bottomrule
\end{tabular}}
\par\vspace{4pt}
\parbox{\linewidth}{\footnotesize
$\Delta$TCR is the paired difference from Endpoint-free; $\Delta$ (Partial $-$ Full) rows give paired Partial $-$ Full differences in each column, computed from unrounded per-video values. $^{*}$Significant; intervals in Tables~\ref{tab:main_ci} and~\ref{tab:endpoint_tcr}; Partial $-$ Full TC-Score intervals are $[-0.028, -0.005]$, $[-0.017, +0.015]$, and $[+0.012, +0.031]$ for Wan2.2, SANA, and Hunyuan. Bold and underline mark the best and second-best value within each backbone block. TCR/DD are percentages (differences in pp); HC is multiplied by 100. HC, BMR, and DD are temporal diagnostics: HC and BMR measure discontinuity and background reversal directly, while DD measures movement, so a drop relative to Endpoint-free indicates motion suppression.}
\par\vspace{-4pt}
\end{table}

\runinsubsection{Endpoint-Free vs.\ Full Conditioning}{sec:full}
For Endpoint-free$\rightarrow$Full (Table~\ref{tab:main}), TCR rises significantly by 8.16 percentage points (pp) on Wan2.2 and 12.23~pp on SANA. DD decreases by 12.62 and 9.90~pp, respectively, while HC and BMR increase on both. Full supplies useful target information while inducing temporal artifacts.

Hunyuan's principal artifacts are discontinuity and background reversal: HC rises from 3.47 to 12.96 and BMR from 0.305 to 0.585, while DD is 51.5\% for Endpoint-free and 52.6\% for Full. Thus, Full's temporal artifacts differ across settings. These observations are consistent with spatial over-specification: outside $\Mzero$, the first--last-frame difference is 0.13--0.23 for Endpoint-free videos but 0.06--0.08 under Full, which returns preserved context near its source-time layout (Appendix~\ref{app:layout}).

Beyond these backbones, FLUX.2 Full conditioning significantly raises TCR on Wan2.2 Base (31.07\% to 45.24\%, $+14.17$~pp) and MfM (32.62\% to 43.88\%, $+11.26$~pp), whereas the gains of OmniWeaving (39.42\% to 43.30\%, $+3.88$~pp) and of Wan2.2 with Qwen-Image-Edit endpoints (33.79\% to 35.15\%, $+1.36$~pp) are not significant (Table~\ref{tab:endpoint_tcr}). An edited endpoint can aid completion across conditioning approaches, but the magnitude and certainty of the gain vary across models and editors. Full's temporal artifacts also persist under learned endpoint conditioning: Full sharply lowers DD by 20.19~pp for MfM and 17.67~pp for OmniWeaving, and raises OmniWeaving's BMR (Table~\ref{tab:learned}).

\begin{figure}[t]
\centering
\includegraphics[width=\linewidth,trim=0 4.8pt 0 3.5pt,clip]{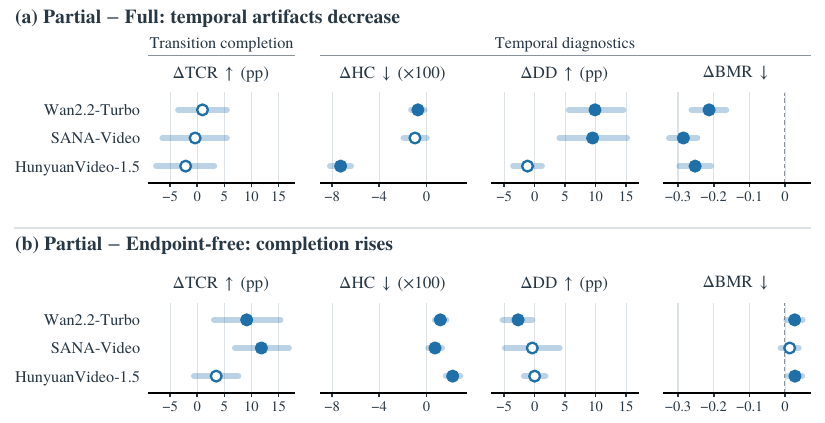}
\caption{\textbf{Partial reduces temporal artifacts relative to Full.} Paired differences with 95\% CIs over 103 prompts and five seeds; open markers are not significant. Tables~\ref{tab:main} and~\ref{tab:main_ci} give levels and intervals.}
\label{fig:effects}
\end{figure}

\runinsubsection{Partial vs.\ Endpoint-Free and Full}{sec:partial}
\textbf{Relative to Full, Partial mitigates temporal artifacts at comparable TCR.} Across Wan2.2, SANA, and Hunyuan, TCR stays within 2.14~pp of Full. HC, DD, and BMR each move toward their Endpoint-free levels, and every significant temporal difference favors Partial (Fig.~\ref{fig:effects}; Tables~\ref{tab:main} and~\ref{tab:main_ci}). The BMR reduction holds within each edit type (Table~\ref{tab:subgroups_temporal}). Outside $\Mzero$, Partial leaves the first--last-frame difference of preserved context within 0.02 of Endpoint-free (Appendix~\ref{app:layout}).

\textbf{The mitigation is not a return to Endpoint-free.} Partial keeps a significant TCR gain over Endpoint-free on Wan2.2 and SANA ($+9.13$ and $+11.84$~pp), and on Wan2.2, global softening at comparable TCR does not reproduce these reductions (Section~\ref{sec:controls}). Some temporal artifacts remain relative to Endpoint-free (Table~\ref{tab:main}). Edit-type breakdowns are in Table~\ref{tab:subgroups}.

\section{Controlled Analysis}
\label{sec:analysis}

Section~\ref{sec:hunyuan} localizes Hunyuan's Full discontinuity at the terminal boundary, and Section~\ref{sec:context} uses a donor-context intervention to test whether terminal content, not only constraint extent, drives Full's temporal artifacts. We run the intervention on Hunyuan and on Wan2.2. Wan2.2's Full artifacts are mainly motion suppression and background reversal, so the Wan2.2 run tests whether the donor effect extends beyond a boundary discontinuity. On Hunyuan, where the discontinuity arises at this boundary, cross-seed donors test how much of the effect depends on matching the recipient trajectory. We use Wan2.2 for global softening, support area, and placement controls because its shared conditioning path applies the same weight ($w=1$, $\alpha$, or $M_0$) to model input and clean-prediction projection, allowing these factors to vary within one mechanism. Section~\ref{sec:fg} then applies the same support to gradient-based guidance on another backbone.

\runinsubsection{Localization of Terminal Discontinuity}{sec:hunyuan}
Unlike Wan2.2 and SANA, Hunyuan Full sharply increases HC without reducing DD (Section~\ref{sec:full}), and the discontinuity is concentrated at the terminal boundary: the adjacent-frame pair entering Hunyuan's terminal latent slot is the minimum-similarity pair in 95.9\% of Full videos, versus 1.7\% for Endpoint-free (Appendix~\ref{app:hc_localization}). Full therefore preserves ordinary motion while introducing a localized discontinuity where the endpoint is imposed. If this discontinuity depends on what the endpoint imposes, not only on where, replacing terminal content at fixed support should reduce it.

\begin{figure}[t]
\centering
\includegraphics[width=\linewidth,trim=0 4.5pt 0 3.5pt,clip]{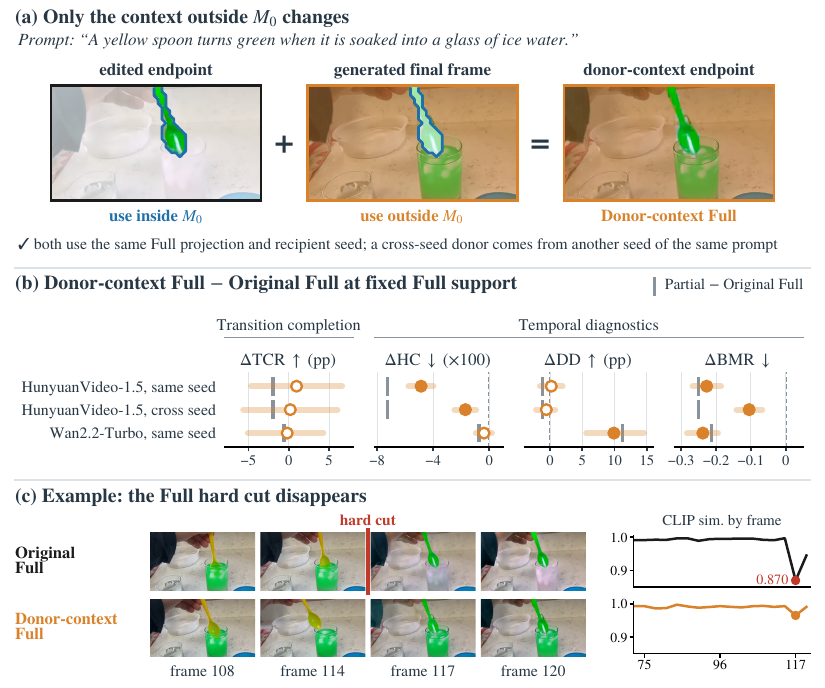}
\caption{\textbf{Terminal content, not only constraint extent, affects temporal artifacts.} (a) The donor-context endpoint keeps the edited endpoint inside $M_0$ (blue outline) and uses the final frame of a paired Partial rollout outside it (orange tint). Original and Donor-context Full share the Full projection and the recipient seed; a cross-seed donor comes from another seed of the same prompt. (b) Donor-context minus Original Full with paired 95\% CIs over 103 prompts and five seeds; open markers are not significant; gray ticks mark Partial minus Original Full. (c) HunyuanVideo-1.5, prompt 28, seed 22004: native frames and consecutive-frame CLIP similarity; the Original Full discontinuity between frames 114 and 117 is absent under Donor-context Full.}
\label{fig:context}
\end{figure}

\runinsubsection{Donor Context under Fixed Full Support}{sec:context}
Full and Partial differ in both imposed content and constraint extent, so we vary terminal content alone. Original Full (O) conditions on the endpoint latent $z_E$. A paired Partial rollout (P) with the same source, prompt, and seed supplies its final RGB frame $I_T^P$. Donor-context Full (D) combines original content inside estimated support with donor context outside:
\begin{equation}
z_D=\Mzero\odot z_E+(1-\Mzero)\odot E(I_T^P),
\label{eq:future_context}
\end{equation}
where $E$ is the backbone's terminal-frame encoding, also used for $z_E$ (Appendix~\ref{app:context}). Here $\Mzero$ selects the source of terminal content, not projection support. Within each backbone, O and D constrain all terminal spatial positions with the same Full support, projection, sampler, source, prompt, and paired noises. The donor requires an extra rollout; it is not ground-truth future context and need not fit the new Full rollout.

\par\noindent\textbf{Terminal content affects temporal artifacts at fixed extent.}
On Hunyuan, D significantly lowers HC from 12.96 to 8.10 and BMR from 0.585 to 0.358 (paired differences $-4.86$ and $-0.227$; relative reductions of 37.5\% and 38.8\%), while TCR and DD remain comparable. On Wan2.2, D significantly lowers BMR by 0.238 and raises DD by 9.90~pp, while TCR and HC remain comparable; TC-Score decreases slightly (Fig.~\ref{fig:context}b; Appendix~\ref{app:context_tc}). Replacing terminal context at fixed Full support thus affects more than Hunyuan's boundary discontinuity.

\par\noindent\textbf{Cross-seed donors.}
To test whether compatibility with the same seed alone explains the donor effect, we take each Hunyuan donor from another seed of the same prompt under an assignment fixed before generation, keeping the recipient seed and its noises (Appendix~\ref{app:context_cross}). Cross-seed donors still significantly lower HC by 1.70 and BMR by 0.105, with TCR and DD comparable, but these reductions are significantly smaller than with same-seed donors (Fig.~\ref{fig:context}b). The improvement therefore does not depend entirely on same-seed pairing.

\par\noindent\textbf{Interpretation.}
Constraint extent alone does not account for the observed temporal artifacts; terminal content also matters. Donor position/visibility, generated-image properties, and trajectory compatibility change jointly. The larger effect with same-seed donors is consistent with a contribution from compatibility with the recipient trajectory. A donor from another seed lacks this compatibility and can even conflict with the recipient's camera and background motion, so the cross-seed effect is a conservative estimate of the role of terminal content. The separate contributions of these three factors remain unidentified. Appendix~\ref{app:context} gives secondary donor-relative and native-latent contrasts. We next examine which original endpoint constraints to retain without a donor rollout.

\runinsubsection{Global Softening vs.\ Spatial Selection}{sec:controls}
Global softening tests whether weakening every terminal constraint can reproduce Partial's reductions in temporal artifacts; spatial controls separately vary how many and which positions are constrained. All comparisons are paired with the main Partial, and all conditions retain the original edited endpoint.

\begin{table}[t]
\centering
\setlength{\tabcolsep}{4pt}
\caption{Selected Wan2.2-Turbo controls: paired comparisons with main Partial (TCR 42.91\%). $A$ is original support area. All conditions and intervals: Tables~\ref{tab:soft}--\ref{tab:support_ci}.}
\label{tab:control_summary}
\fitbox{%
\begin{tabular*}{0.9\linewidth}{@{\extracolsep{\fill}}lr<{\kern6pt}ccc@{}}
\toprule
 & \multicolumn{1}{c}{\kern18pt Transition completion\kern-18pt} & \multicolumn{3}{c}{Temporal diagnostics} \\
\cmidrule(l{22.5pt}r{-13.5pt}){2-2} \cmidrule(l){3-5}
Variant & $\Delta$TCR $\uparrow$ & $\Delta$HC $\downarrow$ & $\Delta$DD $\uparrow$ & $\Delta$BMR $\downarrow$ \\
\midrule
\multicolumn{5}{@{}l}{\textbf{Global softening: Partial $-$ Soft}} \\
Soft ($\alpha=0.75$) & $+0.58\hphantom{{}^{*}}$ & $-1.49^{*}$ & $+7.38^{*}$ & $-0.224^{*}$ \\
\midrule
\multicolumn{5}{@{}l}{\textbf{Spatial controls: Variant $-$ Partial}} \\
$0.5A$ & $-3.50\hphantom{{}^{*}}$ & $-0.53^{*}$ & $+0.39\hphantom{{}^{*}}$ & $-0.017^{*}$ \\
$1.5A$ & $-0.97\hphantom{{}^{*}}$ & $+0.00\hphantom{{}^{*}}$ & $+0.19\hphantom{{}^{*}}$ & $+0.036^{*}$ \\
Shifted $M_0$ & $-6.60^{*}$ & $-0.44^{*}$ & $+0.58\hphantom{{}^{*}}$ & $+0.041^{*}$ \\
Random $A$ \#1 & $-5.24^{*}$ & $-0.04\hphantom{{}^{*}}$ & $-1.36\hphantom{{}^{*}}$ & $+0.028^{*}$ \\
\bottomrule
\end{tabular*}}
\end{table}

At $\alpha=0.75$, TCR is comparable ($+0.58$~pp), but Partial yields significantly higher DD and lower HC and BMR (Table~\ref{tab:control_summary}). The same directions hold at $\alpha=0.5$ and $0.25$ (Tables~\ref{tab:soft} and~\ref{tab:soft_ci}). The tested global softening therefore does not reproduce these reductions.

\runinsubsection{Support Area Analysis}{sec:area}
For original support area $A=\sum_u M_0(u)$, we test $0.5A$, $0.75A$, $1.25A$, and $1.5A$. Distance-based erosion retains interior cells, while expansion adds nearby exterior cells. This varies constraint extent around the same edit-relevant support, with the endpoint and projection fixed.

Reducing area to $0.5A$ or $0.75A$ lowers HC, and $0.5A$ also lowers BMR, while TCR remains comparable. Enlarging to $1.25A$ or $1.5A$ increases BMR, consistent with constraining more preserved context; $1.25A$ also lowers TCR (Tables~\ref{tab:control_summary}, \ref{tab:support}, and~\ref{tab:support_ci}). No tested area is best on every metric.

\runinsubsection{Support Placement Analysis}{sec:placement}
To test whether edit-relevant placement matters beyond area, we relocate $M_0$ away from its original position while preserving its cell count, and compare three random supports targeting area $A$. Construction details are in Appendix~\ref{app:controls}.

At matched area, shifted support lowers TCR by 6.60~pp relative to edit-relevant placement, and all three random supports also lower it significantly (Tables~\ref{tab:control_summary} and~\ref{tab:support}). Shifted support lowers HC while increasing BMR, and DD remains comparable for shifted and random supports. Completion-favorable placement thus need not minimize temporal diagnostics.

\par\noindent\textbf{Overlap with $M_0$ drives completion.}
Random and shifted supports, placed without reference to $M_0$, still cover part of it by chance (9\% on average for random supports), and the completion they retain grows with this coverage (Appendix~\ref{app:placement}). When we control the overlap directly, a support that misses $M_0$ lowers TCR by 12.62~pp, to below Endpoint-free, while one covering 25--75\% of $M_0$ retains about half of the endpoint's completion gain. Random supports of four shapes all fall below Partial (Fig.~\ref{fig:placement}).

\begin{table}[t]
\centering
\setlength{\tabcolsep}{4pt}
\caption{\textbf{Edit-relevant support under gradient-based guidance.} Frame Guidance on CogVideoX-5B-I2V, 103 prompts $\times$ five paired seeds. FG-Full guides the whole last frame; FG-$M_0$ weights the same frame loss by $M_0$. $^{*}$Significant; intervals in Table~\ref{tab:fg}.}
\label{tab:fg_main}
\fitbox{%
\begin{tabular}{@{}lrrrr@{}}
\toprule
 & \multicolumn{1}{c}{Transition completion} & \multicolumn{3}{c}{Temporal diagnostics} \\
\cmidrule(lr){2-2} \cmidrule(l){3-5}
Condition & TCR $\uparrow$ & HC $\downarrow$ & DD $\uparrow$ & BMR $\downarrow$ \\
\midrule
Endpoint-free & 27.18 & 5.44 & 44.7 & 0.388 \\
FG-Full & 36.89 & 5.40 & 30.5 & 0.434 \\
FG-$M_0$ & 37.28 & 6.90 & 42.7 & 0.427 \\
\midrule
FG-Full $-$ Endpoint-free & $+9.71^{*}$ & $-0.04\hphantom{{}^{*}}$ & $-14.17^{*}$ & $+0.045\hphantom{{}^{*}}$ \\
\rowcolor{ForestGreen!12}FG-$M_0$ $-$ FG-Full & $+0.39\hphantom{{}^{*}}$ & $+1.50^{*}$ & $+12.23^{*}$ & $-0.006\hphantom{{}^{*}}$ \\
\bottomrule
\end{tabular}}
\end{table}

\runinsubsection{Gradient-Based Guidance}{sec:fg}
Partial imposes the endpoint by projection, so we test whether restricting the terminal constraint to edit-relevant support also helps under a different mechanism. Frame Guidance steers sampling with a frame loss backpropagated through the video decoder \citep{frameguidance}. We run Frame Guidance's official keyframe configuration on its primary backbone, CogVideoX-5B-I2V \citep{cogvideox}, with the same prompts, seeds, FLUX.2 endpoints, and $\Mzero$; images and $\Mzero$ are center-cropped to 3:2. We compare Endpoint-free, guidance on the whole last frame (FG-Full), and the same loss weighted by $\Mzero$ (FG-$\Mzero$; Table~\ref{tab:fg_main} and Appendix~\ref{app:fg}).

Full-frame guidance reproduces the trade-off of full-frame projection: it raises TCR by 9.71~pp but lowers DD by 14.17~pp. Restricting the loss to $\Mzero$ keeps the completion gain ($+0.39$~pp relative to FG-Full) and raises DD by 12.23~pp, back to the Endpoint-free level, so the benefit of edit-relevant support does not depend on projection. FG-Full barely raises BMR on this backbone, leaving little reversal to remove, and FG-$\Mzero$ raises HC by 1.50 through scene changes early in the video, not through jumps toward the endpoint (Appendix~\ref{app:fg}). Frame Guidance takes about 57 minutes per video on one A6000, against 7 minutes for Endpoint-free; projection adds no backward pass.

\runinsubsection{Diagnostic Cross-Checks}{sec:validation}
\textbf{Discontinuity.} Against blinded Qwen3.6-35B-A3B judgments, HC has AUROC 0.777 (95\% CI $[0.70,0.84]$) on 1,619 evaluable videos from nine generation conditions, excluding Hunyuan.\footnote{Official model card: \href{https://huggingface.co/Qwen/Qwen3.6-35B-A3B}{Qwen3.6-35B-A3B}.} The judge sees all native frames with method identity hidden and identifies abrupt changes lacking a continuous transition. On the Wan2.2 validation subset, judged event incidence is 0.0\% for Endpoint-free, 8.3\% for Full, and 5.0\% for Partial, supporting the direction of the metric comparison. Protocol and sensitivity checks are reported in Appendix~\ref{app:validation}.

\par\noindent\textbf{Reversal and return.}
On Wan2.2, CoTracker3-based Track Reversal Ratio (TRR) and Back-Tracking Ratio (BTR) cross-check BMR \citep{cotracker3}. TRR measures late motion opposing an early direction along tracked trajectories; BTR measures return from the largest achieved displacement to the last visible position. On the common 398/515 prompt--seed pairs meeting tracking and motion criteria under all three conditions, all three diagnostics follow Full $>$ Partial $>$ Endpoint-free (Table~\ref{tab:tracks}). On these pairs, BMR's pooled Spearman correlations with TRR and BTR are 0.64 and 0.65 (descriptive; paired observations are dependent). Main BMR uses all 515 samples per condition (Appendix~\ref{app:validation}). Whole-frame and inside-$\Mzero$ BMR, and BMR without near-static clips, give the same Partial--Full and donor-context conclusions (Appendix~\ref{app:bmr_regions}).

\section{Discussion and Limitations}
\label{sec:discussion}

\textbf{Benefits and limits of partial goals.} In the evaluated settings, imposing only edit-relevant support reduces discontinuity and background reversal and alleviates motion suppression relative to Full at comparable TCR, which global softening does not reproduce. Partial also has specific limits: it does not raise TCR over Full, retains higher HC than Endpoint-free, raises BMR as support grows, and depends on edit-relevant placement (Sections~\ref{sec:partial}--\ref{sec:placement}).

\begin{figure}[t]
\centering
\includegraphics[width=\linewidth,trim=0 7.5pt 0 3pt,clip]{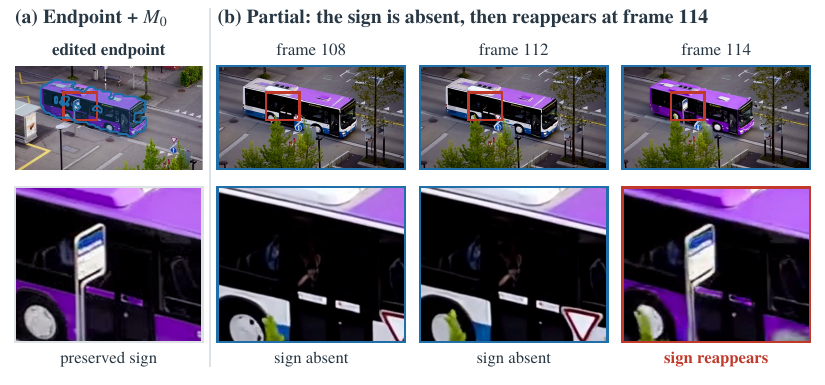}
\caption{\textbf{Estimated edit-relevant support does not guarantee a future-valid terminal constraint.} Wan2.2-Turbo, purple-bus example of Fig.~\ref{fig:support_overview} (case 90001, outside TC-Bench), seed 22001. (a) Saved $M_0$ outline (blue) and token-pooled weights (tint). (b) Partial: the boxed sign is absent at native frames 108 and 112, then reappears at 114. Identical regions are enlarged below.}
\label{fig:fixed_failure}
\end{figure}

\textbf{Future validity.} Edit relevance is a useful support heuristic, not a future-validity estimator. In Fig.~\ref{fig:fixed_failure}, an endpoint-preserved sign reappears under Partial after being absent in earlier frames, illustrating how retained position and visibility constraints can become inappropriate.

\textbf{Generalization and evaluation scope.} We primarily use FLUX.2 endpoints across three backbones, as FLUX.2 scored highest on TCR and TC-Score among tested editors (Table~\ref{tab:editor}; Fig.~\ref{fig:editor_examples}). Donor-context tests cover Hunyuan and Wan2.2; cross-seed donors only Hunyuan, and spatial controls only Wan2.2. Partial remains untested with learned endpoint conditioners, but its support transfers to gradient-based Frame Guidance on a fourth backbone, where it keeps the completion gain and removes motion suppression (Section~\ref{sec:fg}). Diagnostics measure specific temporal artifacts rather than overall video quality.

\section{Conclusion}
\label{sec:conclusion}

Edited endpoints preserve unedited context in the source image's layout, and full-frame conditioning imposes that context as a future requirement, a spatial over-specification. Measured with HC, BMR, and DD, the resulting temporal artifacts appear across five video models spanning training-free and learned endpoint conditioning, in forms that differ by model. Replacing only the preserved context under unchanged Full support reduces these artifacts on two backbones, and on Hunyuan, donors from other seeds retain part of this reduction, so constraint extent alone does not account for them. Imposing only edit-relevant support accordingly mitigates them at comparable TCR, which global softening does not reproduce, while support area and placement trade completion against temporal diagnostics. Residual failures expose the gap between edit relevance and future validity: beyond how strongly to condition on an edited endpoint, the design question is which parts of its content should constrain the future.

\section*{Acknowledgements}

This work was partly supported by the KHIDI grant funded by the Korean government (MOHW) [No.RS-2025-02307233, No.RS-2026-25613012], the NRF or IITP grants funded by the Korean government (MSIT) [No.05-26-04-0094, No.RS-2026-25472075, No.RS-2025-02305581, No.RS-2025-25442338, and No.RS-2021-II211343], the ITIP grant funded by the Korean government (MOTIR) [No.RS-2026-25549946], the Research grant from SNU, and the Strategic Hub grant for International Research Collaboration of SNU. Kyungsu Kim is affiliated with the School of Transdisciplinary Innovations, Department of Biomedical Science, Interdisciplinary Program in AI (IPAI) \& Graduate School of AI, Medical Research Center, and AI Institute at SNU.

\bibliography{references}

\begin{thebibliography}{47}
\providecommand{\natexlab}[1]{#1}
\providecommand{\url}[1]{\texttt{#1}}
\expandafter\ifx\csname urlstyle\endcsname\relax
  \providecommand{\doi}[1]{doi: #1}\else
  \providecommand{\doi}{doi: \begingroup \urlstyle{rm}\Url}\fi

\bibitem[Avrahami et~al.(2023)Avrahami, Fried, and Lischinski]{blendedlatent}
Omri Avrahami, Ohad Fried, and Dani Lischinski.
\newblock {Blended Latent Diffusion}.
\newblock \emph{ACM Transactions on Graphics}, 42\penalty0 (4):\penalty0
  149:1--149:11, 2023.
\newblock \doi{10.1145/3592450}.
\newblock URL \url{https://doi.org/10.1145/3592450}.

\bibitem[Bai et~al.(2025)Bai, Chen, Liu, Wang, Ge, Song, et~al.]{qwen25vl}
Shuai Bai, Keqin Chen, Xuejing Liu, Jialin Wang, Wenbin Ge, Sibo Song, et~al.
\newblock {Qwen2.5-VL Technical Report}.
\newblock \emph{arXiv preprint arXiv:2502.13923}, 2025.
\newblock URL \url{https://arxiv.org/abs/2502.13923}.

\bibitem[Cai et~al.(2026)Cai, Wang, Ye, Liu, Liu, Ye, Wang, Wan, Gai, and
  Yue]{videocanvas}
Minghong Cai, Qiulin Wang, Zongli Ye, Wenze Liu, Quande Liu, Weicai Ye, Xintao
  Wang, Pengfei Wan, Kun Gai, and Xiangyu Yue.
\newblock {VideoCanvas: Unified Video Completion from Arbitrary Spatiotemporal
  Patches via In-Context Conditioning}.
\newblock In \emph{IEEE/CVF Conference on Computer Vision and Pattern
  Recognition (CVPR) Findings}, 2026.
\newblock URL \url{https://arxiv.org/abs/2510.08555}.

\bibitem[Cao et~al.(2023)Cao, Wang, Qi, Shan, Qie, and Zheng]{masactrl}
Mingdeng Cao, Xintao Wang, Zhongang Qi, Ying Shan, Xiaohu Qie, and Yinqiang
  Zheng.
\newblock {MasaCtrl: Tuning-Free Mutual Self-Attention Control for Consistent
  Image Synthesis and Editing}.
\newblock In \emph{Proceedings of the IEEE/CVF International Conference on
  Computer Vision (ICCV)}, pages 22560--22570, 2023.
\newblock URL
  \url{https://openaccess.thecvf.com/content/ICCV2023/html/Cao_MasaCtrl_Tuning-Free_Mutual_Self-Attention_Control_for_Consistent_Image_Synthesis_and_ICCV_2023_paper.html}.

\bibitem[Chen et~al.(2026)Chen, Zhao, Yu, Chu, Chen, Yang, Wang, Pan, Zhou,
  Ling, Liu, Yi, Zhang, Li, Chen, Cai, Fidler, Luo, Han, and Xie]{sana}
Junsong Chen, Yuyang Zhao, Jincheng Yu, Ruihang Chu, Junyu Chen, Shuai Yang,
  Xianbang Wang, Yicheng Pan, Daquan Zhou, Huan Ling, Haozhe Liu, Hongwei Yi,
  Hao Zhang, Muyang Li, Yukang Chen, Han Cai, Sanja Fidler, Ping Luo, Song Han,
  and Enze Xie.
\newblock {SANA-Video: Efficient Video Generation with Block Linear Diffusion
  Transformer}.
\newblock In \emph{International Conference on Learning Representations}, pages
  38973--38997, 2026.
\newblock URL
  \url{https://proceedings.iclr.cc/paper_files/paper/2026/file/41b93c59da0d0f835907fd661d419db2-Paper-Conference.pdf}.

\bibitem[Chen et~al.(2024)Chen, Wang, Zhang, Zhuang, Ma, Yu, Wang, Lin, Qiao,
  and Liu]{seine}
Xinyuan Chen, Yaohui Wang, Lingjun Zhang, Shaobin Zhuang, Xin Ma, Jiashuo Yu,
  Yali Wang, Dahua Lin, Yu~Qiao, and Ziwei Liu.
\newblock {SEINE: Short-to-Long Video Diffusion Model for Generative Transition
  and Prediction}.
\newblock In \emph{International Conference on Learning Representations}, pages
  52208--52222, 2024.
\newblock URL
  \url{https://proceedings.iclr.cc/paper_files/paper/2024/file/e54e6eef11f87a874bf1e4551fc6d04e-Paper-Conference.pdf}.

\bibitem[Couairon et~al.(2023)Couairon, Verbeek, Schwenk, and Cord]{diffedit}
Guillaume Couairon, Jakob Verbeek, Holger Schwenk, and Matthieu Cord.
\newblock {DiffEdit: Diffusion-based semantic image editing with mask
  guidance}.
\newblock In \emph{International Conference on Learning Representations}, 2023.
\newblock URL \url{https://openreview.net/forum?id=3lge0p5o-M-}.

\bibitem[Elmoghany et~al.(2026)Elmoghany, Zhao, Shen, Mukherjee, Zhou, Wu, Lai,
  Yoon, Rossi, Rashwan, Mathur, Manjunatha, Dangi, Nguyen, Lipka, Bui, Singh,
  Zhang, Huang, Cho, Wang, Park, Tu, Chen, Eldardiry, Ahmed, Nguyen, Manocha,
  Elhoseiny, and Dernoncourt]{infinitystory}
Mohamed Elmoghany, Liangbing Zhao, Xiaoqian Shen, Subhojyoti Mukherjee, Yang
  Zhou, Gang Wu, Viet~Dac Lai, Seunghyun Yoon, Ryan Rossi, Abdullah Rashwan,
  Puneet Mathur, Varun Manjunatha, Daksh Dangi, Chien Nguyen, Nedim Lipka,
  Trung Bui, Krishna~Kumar Singh, Ruiyi Zhang, Xiaolei Huang, Jaemin Cho,
  Yu~Wang, Namyong Park, Zhengzhong Tu, Hongjie Chen, Hoda Eldardiry, Nesreen
  Ahmed, Thien Nguyen, Dinesh Manocha, Mohamed Elhoseiny, and Franck
  Dernoncourt.
\newblock {InfinityStory: Unlimited Video Generation with World Consistency and
  Character-Aware Shot Transitions}.
\newblock \emph{arXiv preprint arXiv:2603.03646}, 2026.
\newblock URL \url{https://arxiv.org/abs/2603.03646}.

\bibitem[Feng et~al.(2024)Feng, Ding, Xia, Niklaus, Abrevaya, Black, and
  Zhang]{trf}
Haiwen Feng, Zheng Ding, Zhihao Xia, Simon Niklaus, Victoria Abrevaya,
  Michael~J. Black, and Xuaner Zhang.
\newblock {Explorative Inbetweening of Time and Space}.
\newblock In \emph{European Conference on Computer Vision}, pages 378--395,
  2024.
\newblock \doi{10.1007/978-3-031-73229-4_22}.
\newblock URL \url{https://doi.org/10.1007/978-3-031-73229-4_22}.

\bibitem[Feng et~al.(2025)Feng, Li, Saxon, Fu, Chen, and Wang]{tcbench}
Weixi Feng, Jiachen Li, Michael Saxon, Tsu-Jui Fu, Wenhu Chen, and William~Yang
  Wang.
\newblock {{TC}-Bench: Benchmarking Temporal Compositionality in Conditional
  Video Generation}.
\newblock In \emph{Findings of the Association for Computational Linguistics:
  ACL 2025}, pages 4638--4662, 2025.
\newblock \doi{10.18653/v1/2025.findings-acl.241}.
\newblock URL \url{https://aclanthology.org/2025.findings-acl.241/}.

\bibitem[Ge et~al.(2026)Ge, Zhang, Huang, He, Wang, Ma, Song, Liu, and
  Zhang]{salt}
Xingtong Ge, Yi~Zhang, Yushi Huang, Dailan He, Xiahong Wang, Bingqi Ma, Guanglu
  Song, Yu~Liu, and Jun Zhang.
\newblock {SALT: Self-Consistent Distribution Matching with Cache-Aware
  Training for Few-Step Video Generation}.
\newblock In \emph{European Conference on Computer Vision}, 2026.
\newblock URL \url{https://eccv.ecva.net/virtual/2026/poster/3473}.

\bibitem[Hertz et~al.(2023)Hertz, Mokady, Tenenbaum, Aberman, Pritch, and
  Cohen-Or]{prompttoprompt}
Amir Hertz, Ron Mokady, Jay Tenenbaum, Kfir Aberman, Yael Pritch, and Daniel
  Cohen-Or.
\newblock {Prompt-to-Prompt Image Editing with Cross-Attention Control}.
\newblock In \emph{International Conference on Learning Representations}, 2023.
\newblock URL \url{https://openreview.net/forum?id=_CDixzkzeyb}.

\bibitem[Huang et~al.(2024)Huang, He, Yu, Zhang, Si, Jiang, Zhang, Wu, Jin,
  Chanpaisit, Wang, Chen, Wang, Lin, Qiao, and Liu]{vbench}
Ziqi Huang, Yinan He, Jiashuo Yu, Fan Zhang, Chenyang Si, Yuming Jiang, Yuanhan
  Zhang, Tianxing Wu, Qingyang Jin, Nattapol Chanpaisit, Yaohui Wang, Xinyuan
  Chen, Limin Wang, Dahua Lin, Yu~Qiao, and Ziwei Liu.
\newblock {VBench: Comprehensive Benchmark Suite for Video Generative Models}.
\newblock In \emph{Proceedings of the IEEE/CVF Conference on Computer Vision
  and Pattern Recognition (CVPR)}, pages 21807--21818, 2024.
\newblock URL
  \url{https://openaccess.thecvf.com/content/CVPR2024/html/Huang_VBench_Comprehensive_Benchmark_Suite_for_Video_Generative_Models_CVPR_2024_paper.html}.

\bibitem[Jang et~al.(2026)Jang, Ki, Jo, Yoon, Kim, Lin, and
  Hwang]{frameguidance}
Sangwon Jang, Taekyung Ki, Jaehyeong Jo, Jaehong Yoon, Soo~Ye Kim, Zhe Lin, and
  Sung~Ju Hwang.
\newblock {Frame Guidance: Training-Free Guidance for Frame-Level Control in
  Video Diffusion Models}.
\newblock In \emph{International Conference on Learning Representations}, pages
  60977--61003, 2026.
\newblock URL
  \url{https://proceedings.iclr.cc/paper_files/paper/2026/file/636f1f78f7096caf126afaa27deb7b53-Paper-Conference.pdf}.

\bibitem[Jiang et~al.(2025)Jiang, Han, Mao, Zhang, Pan, and Liu]{vace}
Zeyinzi Jiang, Zhen Han, Chaojie Mao, Jingfeng Zhang, Yulin Pan, and Yu~Liu.
\newblock {VACE: All-in-One Video Creation and Editing}.
\newblock In \emph{Proceedings of the IEEE/CVF International Conference on
  Computer Vision (ICCV)}, pages 17191--17202, 2025.
\newblock URL \url{https://arxiv.org/abs/2503.07598}.

\bibitem[Jiang et~al.(2026)Jiang, Sun, Zeng, Yang, Zhang, Wu, Cheng, Yu, Yang,
  and Wen]{geditv2}
Zhangqi Jiang, Zheng Sun, Xianfang Zeng, Yufeng Yang, Xuanyang Zhang, Yongliang
  Wu, Wei Cheng, Gang Yu, Xu~Yang, and Bihan Wen.
\newblock {{GEditBench v2}: A Human-Aligned Benchmark for General Image
  Editing}.
\newblock \emph{arXiv preprint arXiv:2603.28547}, 2026.
\newblock URL \url{https://arxiv.org/abs/2603.28547}.

\bibitem[Karaev et~al.(2025)Karaev, Makarov, Wang, Neverova, Vedaldi, and
  Rupprecht]{cotracker3}
Nikita Karaev, Yuri Makarov, Jianyuan Wang, Natalia Neverova, Andrea Vedaldi,
  and Christian Rupprecht.
\newblock {CoTracker3: Simpler and Better Point Tracking by Pseudo-Labelling
  Real Videos}.
\newblock In \emph{Proceedings of the IEEE/CVF International Conference on
  Computer Vision (ICCV)}, pages 6013--6022, 2025.
\newblock URL
  \url{https://openaccess.thecvf.com/content/ICCV2025/html/Karaev_CoTracker3_Simpler_and_Better_Point_Tracking_by_Pseudo-Labelling_Real_Videos_ICCV_2025_paper.html}.

\bibitem[Kim et~al.(2025)Kim, Hyung, and Choo]{ticft}
Kinam Kim, Junha Hyung, and Jaegul Choo.
\newblock {Temporal In-Context Fine-Tuning with Temporal Reasoning for
  Versatile Control of Video Diffusion Models}.
\newblock In \emph{Advances in Neural Information Processing Systems},
  volume~38, pages 74491--74518, 2025.
\newblock \doi{10.52202/085713-2501}.
\newblock URL
  \url{https://proceedings.neurips.cc/paper_files/paper/2025/file/6c03ebd7344bcc0c12c4ba8aa8487c00-Paper-Conference.pdf}.

\bibitem[Ku et~al.(2024)Ku, Wei, Ren, Yang, and Chen]{anyv2v}
Max Ku, Cong Wei, Weiming Ren, Harry Yang, and Wenhu Chen.
\newblock {AnyV2V: A Tuning-Free Framework For Any Video-to-Video Editing
  Tasks}.
\newblock \emph{Transactions on Machine Learning Research}, 2024.
\newblock URL \url{https://openreview.net/forum?id=RFrJCkw2oa}.

\bibitem[Li et~al.(2025)Li, Xie, Wu, Zhang, and Wang]{five}
Minghan Li, Chenxi Xie, Yichen Wu, Lei Zhang, and Mengyu Wang.
\newblock {FiVE-Bench: A Fine-grained Video Editing Benchmark for Evaluating
  Emerging Diffusion and Rectified Flow Models}.
\newblock In \emph{Proceedings of the IEEE/CVF International Conference on
  Computer Vision (ICCV)}, pages 16672--16681, 2025.
\newblock URL
  \url{https://openaccess.thecvf.com/content/ICCV2025/html/Li_FiVE-Bench_A_Fine-grained_Video_Editing_Benchmark_for_Evaluating_Emerging_Diffusion_ICCV_2025_paper.html}.

\bibitem[Li et~al.(2026)Li, Yang, Shi, Feng, Wen, Peng, and Zhang]{mfm}
Ruibin Li, Tao Yang, Yangming Shi, Weiguo Feng, Shilei Wen, Bingyue Peng, and
  Lei Zhang.
\newblock {Many-for-Many: Unify the Training of Multiple Video and Image
  Generation and Manipulation Tasks}.
\newblock In \emph{International Conference on Learning Representations}, pages
  59233--59269, 2026.
\newblock URL
  \url{https://proceedings.iclr.cc/paper_files/paper/2026/file/60f9118a849e8e9a0c67e2a36ad80ebf-Paper-Conference.pdf}.

\bibitem[Litman et~al.(2026)Litman, Liu, Seyb, Milef, Zhou, Marshall, Tulsiani,
  and Leak]{editctrl}
Yehonathan Litman, Shikun Liu, Dario Seyb, Nicholas Milef, Yang Zhou, Carl
  Marshall, Shubham Tulsiani, and Caleb Leak.
\newblock {EditCtrl: Disentangled Local and Global Control for Real-Time
  Generative Video Editing}.
\newblock In \emph{Proceedings of the IEEE/CVF Conference on Computer Vision
  and Pattern Recognition (CVPR)}, pages 8965--8975, 2026.
\newblock URL
  \url{https://openaccess.thecvf.com/content/CVPR2026/html/Litman_EditCtrl_Disentangled_Local_and_Global_Control_for_Real-Time_Generative_Video_CVPR_2026_paper.html}.

\bibitem[Liu et~al.(2025)Liu, Wang, Wang, Liu, Zhang, Lee, Li, Yu, Lin, Kim,
  and Jia]{genprop}
Shaoteng Liu, Tianyu Wang, Jui-Hsien Wang, Qing Liu, Zhifei Zhang, Joon-Young
  Lee, Yijun Li, Bei Yu, Zhe Lin, Soo~Ye Kim, and Jiaya Jia.
\newblock {Generative Video Propagation}.
\newblock In \emph{Proceedings of the IEEE/CVF Conference on Computer Vision
  and Pattern Recognition (CVPR)}, pages 17712--17722, 2025.
\newblock URL
  \url{https://openaccess.thecvf.com/content/CVPR2025/html/Liu_Generative_Video_Propagation_CVPR_2025_paper.html}.

\bibitem[Lu et~al.(2025)Lu, Zhou, Bao, Chen, Li, and Zhu]{dpmsolverpp}
Cheng Lu, Yuhao Zhou, Fan Bao, Jianfei Chen, Chongxuan Li, and Jun Zhu.
\newblock {DPM-Solver++: Fast Solver for Guided Sampling of Diffusion
  Probabilistic Models}.
\newblock \emph{Machine Intelligence Research}, 22\penalty0 (4):\penalty0
  730--751, 2025.
\newblock \doi{10.1007/s11633-025-1562-4}.

\bibitem[Lugmayr et~al.(2022)Lugmayr, Danelljan, Romero, Yu, Timofte, and
  Van~Gool]{repaint}
Andreas Lugmayr, Martin Danelljan, Andres Romero, Fisher Yu, Radu Timofte, and
  Luc Van~Gool.
\newblock {RePaint: Inpainting Using Denoising Diffusion Probabilistic Models}.
\newblock In \emph{Proceedings of the IEEE/CVF Conference on Computer Vision
  and Pattern Recognition (CVPR)}, pages 11461--11471, 2022.
\newblock URL
  \url{https://openaccess.thecvf.com/content/CVPR2022/html/Lugmayr_RePaint_Inpainting_Using_Denoising_Diffusion_Probabilistic_Models_CVPR_2022_paper.html}.

\bibitem[Ma et~al.(2024)Ma, Mahdizadehaghdam, Wu, Fan, Gu, Zhao, Shapira, and
  Xie]{maskint}
Haoyu Ma, Shahin Mahdizadehaghdam, Bichen Wu, Zhipeng Fan, Yuchao Gu, Wenliang
  Zhao, Lior Shapira, and Xiaohui Xie.
\newblock {MaskINT: Video Editing via Interpolative Non-autoregressive Masked
  Transformers}.
\newblock In \emph{Proceedings of the IEEE/CVF Conference on Computer Vision
  and Pattern Recognition (CVPR)}, pages 7403--7412, 2024.
\newblock URL
  \url{https://openaccess.thecvf.com/content/CVPR2024/html/Ma_MaskINT_Video_Editing_via_Interpolative_Non-autoregressive_Masked_Transformers_CVPR_2024_paper.html}.

\bibitem[Mao et~al.(2024)Mao, Chen, Gu, Fang, and Shou]{magedit}
Qi~Mao, Lan Chen, Yuchao Gu, Zhen Fang, and Mike~Zheng Shou.
\newblock {MAG-Edit: Localized Image Editing in Complex Scenarios via
  Mask-Based Attention-Adjusted Guidance}.
\newblock In \emph{Proceedings of the 32nd ACM International Conference on
  Multimedia}, pages 6842--6850, 2024.
\newblock \doi{10.1145/3664647.3680830}.
\newblock URL \url{https://doi.org/10.1145/3664647.3680830}.

\bibitem[Minar et~al.(2026)Minar, Oh, Jeong, and Park]{steadyforcing}
Matiur~Rahman Minar, Seunghun Oh, GangHyeon Jeong, and Unsang Park.
\newblock {Steady-Forcing: Balancing Spatial Persistence and Motion Continuity
  in Long-Horizon Nature Video Diffusion}.
\newblock \emph{arXiv preprint arXiv:2606.14732}, 2026.
\newblock URL \url{https://arxiv.org/abs/2606.14732}.

\bibitem[Ouyang et~al.(2026)Ouyang, Teng, Chen, Zhao, and Chen]{dcarl}
Junyi Ouyang, Wenbin Teng, Gonglin Chen, Yajie Zhao, and Haiwei Chen.
\newblock {DCARL: A Divide-and-Conquer Framework for Autoregressive
  Long-Trajectory Video Generation}.
\newblock In \emph{European Conference on Computer Vision}, 2026.
\newblock URL \url{https://eccv.ecva.net/virtual/2026/poster/4282}.

\bibitem[Ouyang et~al.(2024)Ouyang, Dong, Yang, Si, and Pan]{i2vedit}
Wenqi Ouyang, Yi~Dong, Lei Yang, Jianlou Si, and Xingang Pan.
\newblock {I2VEdit: First-Frame-Guided Video Editing via Image-to-Video
  Diffusion Models}.
\newblock In \emph{SIGGRAPH Asia 2024 Conference Papers}, pages 1--11, 2024.
\newblock URL \url{https://doi.org/10.1145/3680528.3687656}.

\bibitem[Pan et~al.(2026)Pan, Tian, Zhang, Kong, Xiong, Long,
  et~al.]{omniweaving}
Kaihang Pan, Qi~Tian, Jianwei Zhang, Weijie Kong, Jiangfeng Xiong, Yanxin Long,
  et~al.
\newblock {{OmniWeaving}: Towards Unified Video Generation with Free-form
  Composition and Reasoning}.
\newblock \emph{arXiv preprint arXiv:2603.24458}, 2026.
\newblock URL \url{https://arxiv.org/abs/2603.24458}.

\bibitem[Perazzi et~al.(2016)Perazzi, Pont-Tuset, McWilliams, Van~Gool, Gross,
  and Sorkine-Hornung]{davis}
Federico Perazzi, Jordi Pont-Tuset, Brian McWilliams, Luc Van~Gool, Markus
  Gross, and Alexander Sorkine-Hornung.
\newblock {A Benchmark Dataset and Evaluation Methodology for Video Object
  Segmentation}.
\newblock In \emph{Proceedings of the IEEE Conference on Computer Vision and
  Pattern Recognition (CVPR)}, pages 724--732, 2016.
\newblock URL
  \url{https://openaccess.thecvf.com/content_cvpr_2016/html/Perazzi_A_Benchmark_Dataset_CVPR_2016_paper.html}.

\bibitem[Podell et~al.(2024)Podell, English, Lacey, Blattmann, Dockhorn,
  M{\"u}ller, Penna, and Rombach]{sdxl}
Dustin Podell, Zion English, Kyle Lacey, Andreas Blattmann, Tim Dockhorn, Jonas
  M{\"u}ller, Joe Penna, and Robin Rombach.
\newblock {SDXL: Improving Latent Diffusion Models for High-Resolution Image
  Synthesis}.
\newblock In \emph{International Conference on Learning Representations}, pages
  1862--1874, 2024.
\newblock URL
  \url{https://proceedings.iclr.cc/paper_files/paper/2024/file/081b08068e4733ae3e7ad019fe8d172f-Paper-Conference.pdf}.

\bibitem[Radford et~al.(2021)Radford, Kim, Hallacy, Ramesh, Goh, Agarwal,
  Sastry, Askell, Mishkin, Clark, Krueger, and Sutskever]{clip}
Alec Radford, Jong~Wook Kim, Chris Hallacy, Aditya Ramesh, Gabriel Goh,
  Sandhini Agarwal, Girish Sastry, Amanda Askell, Pamela Mishkin, Jack Clark,
  Gretchen Krueger, and Ilya Sutskever.
\newblock {Learning Transferable Visual Models From Natural Language
  Supervision}.
\newblock In \emph{Proceedings of the 38th International Conference on Machine
  Learning}, volume 139, pages 8748--8763, 2021.
\newblock URL \url{https://proceedings.mlr.press/v139/radford21a.html}.

\bibitem[Teed and Deng(2020)]{raft}
Zachary Teed and Jia Deng.
\newblock {RAFT: Recurrent All-Pairs Field Transforms for Optical Flow}.
\newblock In \emph{European Conference on Computer Vision}, 2020.
\newblock URL
  \url{https://www.ecva.net/papers/eccv_2020/papers_ECCV/html/3526_ECCV_2020_paper.php}.

\bibitem[Wan et~al.(2025)Wan, Wang, Ai, Wen, Mao, Xie, et~al.]{wan}
Team Wan, Ang Wang, Baole Ai, Bin Wen, Chaojie Mao, Chen-Wei Xie, et~al.
\newblock {Wan: Open and Advanced Large-Scale Video Generative Models}.
\newblock \emph{arXiv preprint arXiv:2503.20314}, 2025.
\newblock URL \url{https://arxiv.org/abs/2503.20314}.

\bibitem[Wang et~al.(2025)Wang, Zhou, Curless, Kemelmacher-Shlizerman,
  Holynski, and Seitz]{generative_inbetweening}
Xiaojuan Wang, Boyang Zhou, Brian Curless, Ira Kemelmacher-Shlizerman,
  Aleksander Holynski, and Steve Seitz.
\newblock {Generative Inbetweening: Adapting Image-to-Video Models for Keyframe
  Interpolation}.
\newblock In \emph{International Conference on Learning Representations}, pages
  30594--30609, 2025.
\newblock URL
  \url{https://proceedings.iclr.cc/paper_files/paper/2025/file/4bbdef62653d8088717640e7660a1ebb-Paper-Conference.pdf}.

\bibitem[Wang and Deng(2026)]{waft}
Yihan Wang and Jia Deng.
\newblock {WAFT: Warping-Alone Field Transforms for Optical Flow}.
\newblock In \emph{International Conference on Learning Representations}, pages
  157389--157402, 2026.
\newblock URL
  \url{https://proceedings.iclr.cc/paper_files/paper/2026/file/ff2853ed92a6e6983ecccd9c922a42d2-Paper-Conference.pdf}.

\bibitem[Wu et~al.(2025{\natexlab{a}})Wu, Zou, Li, Huang, Yang, Tan,
  et~al.]{hunyuan}
Bing Wu, Chang Zou, Changlin Li, Duojun Huang, Fang Yang, Hao Tan, et~al.
\newblock {HunyuanVideo 1.5 Technical Report}.
\newblock \emph{arXiv preprint arXiv:2511.18870}, 2025{\natexlab{a}}.
\newblock URL \url{https://arxiv.org/abs/2511.18870}.

\bibitem[Wu et~al.(2025{\natexlab{b}})Wu, Li, Zhou, Lin, Gao, Yan, Yin,
  et~al.]{qwenimage}
Chenfei Wu, Jiahao Li, Jingren Zhou, Junyang Lin, Kaiyuan Gao, Kun Yan,
  Sheng-ming Yin, et~al.
\newblock {Qwen-Image Technical Report}.
\newblock \emph{arXiv preprint arXiv:2508.02324}, 2025{\natexlab{b}}.
\newblock URL \url{https://arxiv.org/abs/2508.02324}.

\bibitem[Wu et~al.(2026)Wu, Ren, Shen, Cao, He, Lu, Gao, Xie, Lan, Alvarez,
  Gao, Fidler, Wang, and Ling]{chronoedit}
Jay~Zhangjie Wu, Xuanchi Ren, Tianchang Shen, Tianshi Cao, Kai He, Yifan Lu,
  Ruiyuan Gao, Enze Xie, Shiyi Lan, Jose~M. Alvarez, Jun Gao, Sanja Fidler,
  Zian Wang, and Huan Ling.
\newblock {ChronoEdit: Towards Temporal Reasoning for In-Context Image Editing
  and World Simulation}.
\newblock In \emph{International Conference on Learning Representations}, pages
  9886--9899, 2026.
\newblock URL
  \url{https://proceedings.iclr.cc/paper_files/paper/2026/file/10b99813d5992673da90a499b6bcd5f0-Paper-Conference.pdf}.

\bibitem[Yang et~al.(2025{\natexlab{a}})Yang, Kwon, and Ye]{vibidsampler}
Serin Yang, Taesung Kwon, and Jong~Chul Ye.
\newblock {ViBiDSampler: Enhancing Video Interpolation Using Bidirectional
  Diffusion Sampler}.
\newblock In \emph{International Conference on Learning Representations}, pages
  3441--3459, 2025{\natexlab{a}}.
\newblock URL
  \url{https://proceedings.iclr.cc/paper_files/paper/2025/file/098a3d992f0d2128bab619b3c654e1c3-Paper-Conference.pdf}.

\bibitem[Yang et~al.(2025{\natexlab{b}})Yang, Teng, Zheng, Ding, Huang, Xu,
  Yang, Hong, Zhang, Feng, Yin, Gu, Zhang, Wang, Cheng, Liu, Xu, Dong, and
  Tang]{cogvideox}
Zhuoyi Yang, Jiayan Teng, Wendi Zheng, Ming Ding, Shiyu Huang, Jiazheng Xu,
  Yuanming Yang, Wenyi Hong, Xiaohan Zhang, Guanyu Feng, Da~Yin, Xiaotao Gu,
  Yuxuan Zhang, Weihan Wang, Yean Cheng, Ting Liu, Bin Xu, Yuxiao Dong, and Jie
  Tang.
\newblock {CogVideoX: Text-to-Video Diffusion Models with An Expert
  Transformer}.
\newblock In \emph{International Conference on Learning Representations},
  2025{\natexlab{b}}.

\bibitem[Yang et~al.(2025{\natexlab{c}})Yang, Zhang, Yu, Lu, and Bai]{vtg}
Zuhao Yang, Jiahui Zhang, Yingchen Yu, Shijian Lu, and Song Bai.
\newblock {Versatile Transition Generation with Image-to-Video Diffusion}.
\newblock In \emph{Proceedings of the IEEE/CVF International Conference on
  Computer Vision (ICCV)}, pages 16981--16990, 2025{\natexlab{c}}.
\newblock URL
  \url{https://openaccess.thecvf.com/content/ICCV2025/html/Yang_Versatile_Transition_Generation_with_Image-to-Video_Diffusion_ICCV_2025_paper.html}.

\bibitem[Zhang et~al.(2018)Zhang, Isola, Efros, Shechtman, and Wang]{lpips}
Richard Zhang, Phillip Isola, Alexei~A. Efros, Eli Shechtman, and Oliver Wang.
\newblock {The Unreasonable Effectiveness of Deep Features as a Perceptual
  Metric}.
\newblock In \emph{Proceedings of the IEEE Conference on Computer Vision and
  Pattern Recognition (CVPR)}, pages 586--595, 2018.
\newblock URL
  \url{https://openaccess.thecvf.com/content_cvpr_2018/html/Zhang_The_Unreasonable_Effectiveness_CVPR_2018_paper.html}.

\bibitem[Zhao et~al.(2024)Zhao, Zhu, Xiang, Zheng, Li, and Zhu]{imageleakage}
Min Zhao, Hongzhou Zhu, Chendong Xiang, Kaiwen Zheng, Chongxuan Li, and Jun
  Zhu.
\newblock {Identifying and Solving Conditional Image Leakage in Image-to-Video
  Diffusion Model}.
\newblock In \emph{Advances in Neural Information Processing Systems},
  volume~37, pages 30300--30326, 2024.
\newblock \doi{10.52202/079017-0954}.
\newblock URL
  \url{https://proceedings.neurips.cc/paper_files/paper/2024/file/35cb54b887e7aafe74829677cce6c5c6-Paper-Conference.pdf}.

\bibitem[Zou et~al.(2024)Zou, Tang, Zhou, He, Zhao, Zhang, Hu, and
  Sun]{instdiffedit}
Siyu Zou, Jiji Tang, Yiyi Zhou, Jing He, Chaoyi Zhao, Rongsheng Zhang, Zhipeng
  Hu, and Xiaoshuai Sun.
\newblock {Towards Efficient Diffusion-Based Image Editing with Instant
  Attention Masks}.
\newblock In \emph{Proceedings of the AAAI Conference on Artificial
  Intelligence}, volume~38, pages 7864--7872, 2024.
\newblock \doi{10.1609/aaai.v38i7.28622}.
\newblock URL \url{https://ojs.aaai.org/index.php/AAAI/article/view/28622}.

\end{thebibliography}
\bibliographystyle{plainnat}

\appendix
\clearpage

\renewcommand{\thesection}{\Alph{section}}

\makeatletter
\@addtoreset{table}{section}
\@addtoreset{figure}{section}
\makeatother
\renewcommand{\thetable}{\thesection\arabic{table}}
\renewcommand{\thefigure}{\thesection\arabic{figure}}
\def\theHtable{\thesection.\arabic{table}}
\def\theHfigure{\thesection.\arabic{figure}}
\setcounter{table}{0}
\setcounter{figure}{0}

\makeatletter
\@addtoreset{equation}{section}
\makeatother
\renewcommand{\theequation}{\thesection\arabic{equation}}
\def\theHequation{\thesection.\arabic{equation}}
\setcounter{equation}{0}

\phantomsection
{\huge\centering\bfseries Appendix\par}
\addcontentsline{toc}{section}{Appendix}
\vspace{1.5em}
{\large\centering\bfseries Contents\par}
\vspace{1em}

\startcontents[appendix]
\begingroup
\titlecontents{section}[1.7em]
  {\addvspace{0.4em}\bfseries}
  {\contentslabel{1.7em}}{}
  {\titlerule*[0.7pc]{.}\contentspage}
\titlecontents{subsection}[4.1em]
  {\addvspace{0.1em}}
  {\contentslabel{2.4em}}{}
  {\titlerule*[0.7pc]{.}\contentspage}
\titlecontents{subsubsection}[7.2em]
  {\addvspace{0em}}
  {\contentslabel{3.1em}}{}
  {\titlerule*[0.7pc]{.}\contentspage}
\printcontents[appendix]{}{1}{\setcounter{tocdepth}{3}}
\endgroup
\clearpage

\section{Implementation Details}
\label{app:implementation}

\subsection{Endpoint and Support Construction}

\paragraph{Endpoint generation.}
We generate one endpoint per prompt with FLUX.2-klein-base-9B, seed 42, 50 editing steps, bfloat16 inference, and default guidance. We center-crop the source to the target aspect ratio and resize it with Lanczos to $1280\times704$. We use the following final-frame prompt:

\begin{quote}\small
Use the input image as the starting frame of a video. The following sentence describes a transition that happens over the course of the video: ``\{transition\}''

Generate a plausible final frame of that video: the moment right after the transition described above has fully completed.

The frame should look like a later moment from the same continuous shot.

Preserve scene continuity, identity, camera framing, and all persistent background/content, changing only what the transition describes.
\end{quote}

\begin{figure}[t]
\centering
\includegraphics[width=\linewidth]{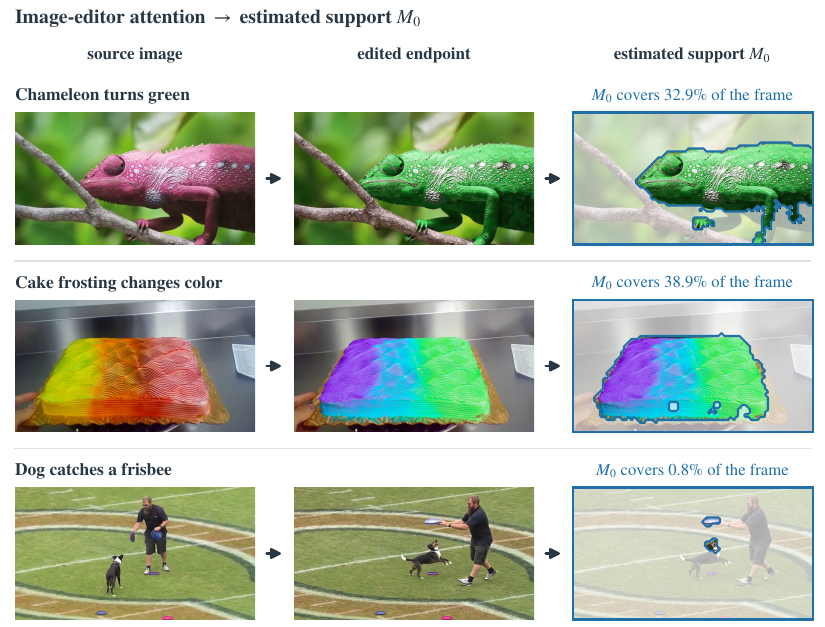}
\caption{\textbf{Estimated edit-relevant support.} Source images, edited endpoints, and GMM--Ising masks for TC-Bench prompts 0, 5, and 82. Blue marks support; percentages give constrained area before token pooling.}
\label{fig:support_examples}
\end{figure}

\paragraph{Attention aggregation.}
Each prompt has one object phrase and one edit phrase, fixed across conditions; they were annotated with the assistance of Claude Opus 5 and manually verified. In each joint-attention layer, the softmax runs over all text, endpoint-image, and reference-image tokens. For a phrase token, we keep the post-softmax entries between that token and the endpoint-image tokens, discard the reference-image entries, and average the text-to-image and image-to-text directions. These maps are averaged over heads, the 32 transformer blocks, and the 50 editing steps of the conditional pass, and then over each phrase's tokens, giving $a_o$ and $a_e$. We combine the normalized maps as

\begin{equation}
a(u)=\tfrac12\left[\mathcal N(a_o)(u)+\mathcal N(a_e)(u)\right],
\qquad
\mathcal N(a)(u)=\frac{a(u)-\min_v a(v)}{\max(\max_v a(v)-\min_v a(v),10^{-12})}.
\end{equation}
We estimate support on the $44\times80$ latent grid (height $\times$ width) using a two-component Gaussian mixture with shared variance:
\begin{equation}
p(a(u))=\sum_{j\in\{0,1\}}\pi_j\mathcal N(a(u);\mu_j,\sigma^2),
\qquad \mu_0<\mu_1.
\end{equation}
We initialize the means at 25\% and 75\% of the observed value range, with equal mixture weights and empirical variance. Variance floors are $10^{-6}$ at initialization and $10^{-8}$ during updates. Expectation--maximization runs for up to 300 iterations, stopping when both means satisfy $|\mu_j^{\rm new}-\mu_j|\le10^{-12}+10^{-5}|\mu_j|$. The high-mean component represents edit-relevant evidence. We encourage spatial coherence with
\begin{equation}
\mathcal E(M)=\sum_u\left[\frac{(a(u)-\mu_{M(u)})^2}{2\sigma^2}-\log\pi_{M(u)}\right]
+\beta\sum_{(u,v)\in\mathcal E_4}\mathbf1[M(u)\ne M(v)].
\label{eq:ising}
\end{equation}
Here $\mathcal E_4$ is the four-neighbor grid and $\beta=2$: in each update, each of a cell's four neighbors with a different label adds $\beta$ to that label's cost. We initialize labels by unary cost and apply up to 12 synchronous conditional-mode updates, replicating border labels as padding, and stop when labels no longer change. Figure~\ref{fig:support_examples} shows the resulting supports.

\subsection{Choice of Endpoint Editor}
\label{app:editor}

Partial needs only an edited endpoint and a support estimated from the editor's text--image attention (Appendix~\ref{app:implementation}), so the primary editor must be open-weight with accessible attention. We use FLUX.2-klein-base-9B for its editing quality, completion, and layout preservation (Table~\ref{tab:editor}).

\paragraph{Editing quality.}
On GEditBench v2, FLUX.2 [klein] 9B ranks first among open-weight editors with an Elo of 1,039, on par with Qwen-Image-Edit-2511 at 1,038 \citep{geditv2}. That evaluation uses the 4-step distilled checkpoint; we use its undistilled base, whose 50-step sampler provides the attention we aggregate.

\paragraph{Completion and temporal diagnostics.}
Under Full conditioning on Wan2.2, FLUX.2 endpoints give the highest TCR of the three editors with the fewest parameters. Paired TCR differences are comparable, $+1.75\;[-3.69, +7.38]$~pp versus ChronoEdit \citep{chronoedit} and $+6.80\;[-0.39, +13.98]$~pp versus Qwen-Image-Edit \citep{qwenimage}, but only FLUX.2 endpoints significantly raise TCR over Endpoint-free (Table~\ref{tab:endpoint_tcr}). TC-Score is significantly higher than with Qwen-Image-Edit endpoints, $+0.035\;[+0.018, +0.052]$, and comparable with ChronoEdit endpoints, $+0.014\;[-0.001, +0.029]$. Full's temporal artifacts are not specific to FLUX.2: relative to Endpoint-free, all three editors significantly raise HC. Qwen-Image-Edit endpoints also change BMR by $+0.256\;[+0.207, +0.307]$ and DD by $-15.34\;[-20.19, -10.68]$~pp, and ChronoEdit endpoints by $+0.198\;[+0.148, +0.249]$ and $-11.46\;[-16.89, -6.02]$~pp. HC, DD, and BMR are comparable between FLUX.2 and Qwen-Image-Edit endpoints.

\paragraph{Layout preservation.}
FLUX.2 endpoints differ from the source least outside $\Mzero$ and most inside it. The outside difference is lower than for Qwen-Image-Edit, $-0.022\;[-0.035, -0.008]$, and ChronoEdit, $-0.038\;[-0.056, -0.023]$. Because $\Mzero$ is estimated from FLUX.2 attention, we also measure Qwen-Image-Edit endpoints outside the support estimated from their own attention. They still differ more from the source than FLUX.2 endpoints do (0.080 versus 0.056; FLUX.2 minus Qwen-Image-Edit, $-0.024\;[-0.038, -0.010]$). Table~\ref{tab:editor} reports 0.077 for Qwen-Image-Edit because it measures outside the FLUX.2 support. FLUX.2 endpoints thus most closely follow Eq.~\eqref{eq:decomposition}, making Full's temporal artifacts less likely to stem from unrequested changes outside the edit.

\begin{figure}[t]
\centering
\includegraphics[width=\linewidth]{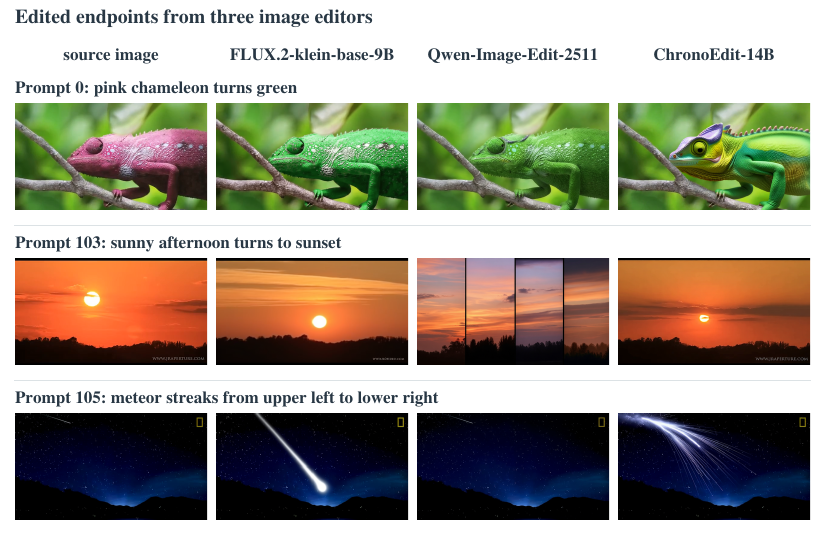}
\caption{\textbf{Endpoints from three editors.} Source images and edited endpoints for TC-Bench prompts 0, 103, and 105. FLUX.2 and Qwen-Image-Edit use the final-frame prompt in Appendix~\ref{app:implementation}; ChronoEdit uses the original transition prompt to match its formulation of editing as video generation.}
\label{fig:editor_examples}
\end{figure}

\paragraph{Qualitative examples.}
In Figure~\ref{fig:editor_examples}, FLUX.2 and Qwen-Image-Edit recolor the chameleon while keeping its pose and texture, whereas ChronoEdit also changes its appearance. For the sunset, FLUX.2 and ChronoEdit lower the sun while keeping the skyline, whereas Qwen-Image-Edit returns a four-panel collage. For the meteor, FLUX.2 extends the streak toward the lower right, Qwen-Image-Edit leaves the source nearly unchanged, and ChronoEdit renders a spray of streaks near the upper left.

\paragraph{Other editors.}
Support estimation uses only the endpoint and the attention to the annotated object and edit phrases. Open-weight diffusion and flow editors expose this attention in joint or cross-attention layers, so we expect the procedure to extend to them without changing the video model. Adapting our attention recorder to Qwen-Image-Edit's token layout, with the same GMM--Ising rule, yields supports for all 103 prompts, covering 6.3\% of the frame on average versus 8.9\% for FLUX.2. We did not evaluate Partial with these supports; editors without attention access, such as closed APIs, would need supports from another source.

\begin{table}[!tbp]
\centering
\caption{Endpoint editors under Full conditioning on Wan2.2-Turbo (103 prompts $\times$ five paired seeds). Out and In give the mean absolute RGB difference between the endpoint and the source image outside and inside the FLUX.2 support $\Mzero$. ChronoEdit uses the original transition prompt and the last frame of a five-frame generation (Appendix~\ref{app:overview}).}
\label{tab:editor}
\setlength{\tabcolsep}{4pt}
\fitbox{%
\begin{tabular}{@{}lrrrrrr@{\hspace{12pt}}*{2}{w{c}{0.09\linewidth}}@{}}
\toprule
 & & \multicolumn{5}{c}{Full on Wan2.2-Turbo} & \multicolumn{2}{c}{Endpoint vs.\ source} \\
\cmidrule(lr){3-7}\cmidrule(l){8-9}
Editor & Params & TCR $\uparrow$ & TC-Score $\uparrow$ & HC $\downarrow$ & DD $\uparrow$ & BMR $\downarrow$ & Out & In \\
\midrule
FLUX.2-klein-base-9B & 9B & 41.94 & 0.8352 & 6.47 & 66.8 & 0.599 & 0.056 & 0.232 \\
Qwen-Image-Edit-2511 & 20B & 35.15 & 0.8001 & 7.11 & 64.1 & 0.617 & 0.077 & 0.161 \\
ChronoEdit-14B & 14B & 40.19 & 0.8214 & 6.13 & 68.0 & 0.558 & 0.094 & 0.192 \\
\midrule
Endpoint-free & -- & 33.79 & 0.7841 & 4.57 & 79.4 & 0.360 & -- & -- \\
\bottomrule
\end{tabular}}
\end{table}

\subsection{Video Configurations and Projection}

\begin{table}[!tbp]
\centering
\caption{Primary generation configurations. Spatial dimensions are width by height. \mbox{HunyuanVideo-1.5} selects an aspect-ratio-compatible resolution using its 480p configuration.}
\label{tab:config}
\fitbox{%
\begin{tabular}{lrrrrl}
\toprule
Backbone & Steps & Frames & FPS & Resolution & Terminal condition \\
\midrule
Wan2.2-Turbo & 4 & 121 & 24 & $1280\times704$ & Input/clean projection \\
SANA-Video 2B & 50 & 81 & 16 & $1280\times704$ & See wrappers below \\
HunyuanVideo-1.5 & 8 & 121 & 24 & 480p & Noise-matched projection \\
\bottomrule
\end{tabular}}
\end{table}

We use seeds 22001--22005 and share each edited endpoint between Full and Partial. Wan2.2 uses the distilled TI2V-5B checkpoint; SANA uses the 2B 720p checkpoint with guidance scale 6.0.

Wan2.2 average-pools $M_0$ in $2\times2$ blocks from the $44\times80$ latent grid to a $22\times40$ token grid and repeats the weights over the corresponding latent cells. SANA area-pools to its $22\times40$ latent/token grid. Pooled weights can be fractional; they are broadcast over latent channels, with token timesteps scaled by $1-w$. At each step, the terminal state $x$ enters the Wan2.2 model as $x_{\rm in}=w\odot z_E+(1-w)\odot x$, and the clean prediction is projected with Eq.~\eqref{eq:partial} before the sampling update; both blends use the same $w$, including $w=\alpha$ for Soft.

SANA Partial keeps the solver state $x$, which starts from the initial noise like every other latent slot, separate from the blended model input $y$. For the terminal slot at noise level $\sigma$ and timestep $t$,
\begin{equation}
y=w\odot z_E+(1-w)\odot x,\qquad
\hat z=w\odot z_E+(1-w)\odot(y-\sigma v),\qquad
\tilde v=(x-\hat z)/\sigma,
\label{eq:sana}
\end{equation}
where $v$ is the model's flow for input $y$ at token timesteps $t(1-w)$. The corrected flow $\tilde v$ replaces $v$ wherever $w>0$, and the native DPM-Solver++ update advances $x$ \citep{dpmsolverpp}; the first latent slot stays clean and outside the solver. SANA Full instead clamps the terminal state at initialization and after each scheduler step, with zero terminal token timesteps. With $M_0\equiv1$, the Partial update reduces to this clamp: from identical initial noise, the two implementations' final latents differ by at most $1.2\times10^{-7}$ in two verification videos.

For Hunyuan, we resize and center-crop the source and mask to the selected 480p aspect ratio, then pool and threshold the mask at 0.5. We repeat the endpoint over five frames, encode it with the causal video VAE, and select the second temporal latent slot, which encodes four copies of the endpoint just as the terminal slot encodes four frames. Endpoint noise $\epsilon_E$ is sampled once from a seed derived from the video seed and held fixed. At noise level $\sigma_k$, masked projection targets
\begin{equation}
z_E^{(k)}=(1-\sigma_k)z_E+\sigma_k\epsilon_E,
\label{eq:noised}
\end{equation}
at initialization and after each scheduler update, ending with clean projection at zero noise.

Spatial controls are paired with the main Partial. Additional completion comparisons use learned endpoint conditioning in MfM and OmniWeaving, and 50 sampling steps for Wan2.2 Base.

\subsection{Support-Size and Placement Controls}
\label{app:controls}

For $A=\sum_u M_0(u)$, size controls target rounded areas $0.5A$, $0.75A$, $1.25A$, and $1.5A$. Euclidean distance transforms rank cells: erosion retains the deepest interior cells, while expansion adds the nearest exterior cells. Seeded random values break ties.

Shifted support preserves area while changing location. On the $44\times80$ latent grid, we consider translations that keep the mask inside the frame and use even row and column offsets, which keep $2\times2$ token blocks aligned; we sample uniformly among those with IoU below 0.1, or take the lowest-overlap candidate. When no such translation exists, we shift the mask by half the frame width, with a vertical shift of zero or half the frame height, clip it at the border, and regrow it to its original cell count with the distance-based expansion of the size controls, keeping the candidate with the lowest IoU; this occurs for 12 of 103 prompts.

Random support selects $\operatorname{round}(A/4)$ cells on the $22\times40$ token grid and expands each to a $2\times2$ latent block. Area therefore matches $M_0$ to within two latent cells. We evaluate three independent random supports per prompt. These supports remain fixed during sampling; random release in Appendix~\ref{app:failures} instead shrinks support over denoising steps.

\FloatBarrier
\section{Metric Definitions and Evaluation Populations}
\label{app:metrics}

\paragraph{Temporal sampling and preprocessing.}
For $N$ native frames at FPS $f$, TC-Score uses unique indices $\operatorname{round}(k(N-1)/15)$, $k=0,\ldots,15$. HC samples $\operatorname{round}(kf/8)$ for $k=0,\ldots,\lfloor 8(N-1)/f\rfloor$; DD and BMR take every $\max(1,\operatorname{round}(f/8))$-th frame. Every evaluated video has $f\in\{8,16,24\}$, so all three temporal diagnostics select the same frames at 8 FPS. CLIP uses its standard 224-pixel image preprocessing; flow uses decoded resolution. Track diagnostics use native 24 FPS frames resized to $640\times352$.

\subsection{Transition Completion and TC-Score}

We evaluate 50 attribute-binding, 38 object-binding, and 15 background-shift prompts using the benchmark assertions and Qwen2.5-VL-32B-Instruct. TCR requires Yes for every assertion except those about ``other objects.''

TC-Score combines a VLM assertion average, excluding transition-object consistency, with mean adjacent-frame CLIP ViT-B/32 similarity on the 16-frame grid. We map CLIP similarity from $[0.9,0.98]$ to $[0,1]$ and clip it. The VLM and CLIP terms receive weights $2/3$ and $1/3$, respectively. Adjacent-frame CLIP similarity falls with larger motion and discontinuities, so the CLIP term mixes completion with properties that DD and HC measure, and HC uses the same similarity (Appendix~\ref{app:context_tc}). We therefore treat TCR as the primary completion measure and report TC-Score for comparison with TC-Bench.

\subsection{Dynamic Degree and Hard-Cut}

DD follows VBench. For each adjacent pair, $m(t)$ is the mean of the largest 5\% of RAFT flow magnitudes. A video is dynamic if $m(t)>6\min(H,W)/256$ for at least $\operatorname{round}(4N/16)$ pairs, where $N$ is the sampled frame count. The main clips have 41 frames at 8 FPS and require 10 qualifying pairs. DD measures substantial movement, including camera motion.

HC is the difference between the median and minimum adjacent-frame CLIP cosine similarities (Eq.~\eqref{eq:hc}), an absolute rather than proportional drop. We report $100\times$HC, including its confidence intervals. The minimum-similarity pair $t_{\rm HC}$ identifies the event location.

\subsection{Background Motion Reversal}

BMR uses WAFT flow at 8 FPS outside the FLUX.2-derived support. We keep this measurement mask fixed across all conditions, resize it with nearest-neighbor interpolation, and take a deterministic strided subset of at most 80,000 pixels. Equation~\eqref{eq:bmr} weights flow by magnitude to prevent static pixels from dominating the direction estimate.

The reference axis is the normalized componentwise median flow over the first half of the video; reversal is measured over the second half. The ratio is undefined for a zero reference or zero late along-axis travel. All main and control conditions have 515 finite values, without DD filtering. We cross-check BMR with track diagnostics in Appendix~\ref{app:tracks} and repeat the main comparisons over other measurement regions and without near-static clips in Appendix~\ref{app:bmr_regions}.

\subsection{Evaluation Populations}
\label{app:populations}

Table~\ref{tab:populations} summarizes the cohorts used in each analysis.

\begin{table}[ht]
\centering
\renewcommand{\arraystretch}{1.1}
\caption{Appendix populations and references. Counts refer to paired prompt--seed positions unless stated otherwise.}
\label{tab:populations}
\begin{tabularx}{\linewidth}{@{}>{\raggedright\arraybackslash}p{0.28\linewidth}>{\raggedright\arraybackslash}X@{}}
\toprule
Analysis & Population and reference \\
\midrule
Main backbone comparisons & 103 prompts $\times$ 5 seeds per condition; Table~\ref{tab:main}. \\
\addlinespace[7pt]
Wan2.2-Turbo spatial controls & 103 prompts $\times$ 5 seeds per support, paired with the main Wan2.2-Turbo Partial (TCR 42.91\%). \\
\addlinespace[7pt]
HunyuanVideo-1.5 donor-context intervention & Original Full reuses the main cohort. Donor P regenerates main Partial with identical settings; TCR 44.66\% versus 44.47\% reflects run-to-run variation. \\
\addlinespace[7pt]
HunyuanVideo-1.5 cross-seed donors & O, same-seed D, and P are reused; 515 new $D_{\rm cross}$ videos use a donor-seed assignment fixed before generation. \\
\addlinespace[7pt]
Wan2.2-Turbo donor-context intervention & O, D, N, and P are new 515-video runs sharing one noise stream per prompt--seed. P reproduces main Partial's HC, DD, and BMR exactly; TCR 42.72\% versus 42.91\% reflects re-judging variation. O regenerates Full with this noise stream and stays within sampling variation of main Full on all four metrics (TCR 43.30\% versus 41.94\%, HC 6.49 versus 6.47, DD 65.44\% versus 66.80\%, BMR 0.600 versus 0.599; no paired difference is significant). \\
\addlinespace[7pt]
HC model-judge agreement & 90 prompts $\times$ 2 samples $\times$ 9 conditions; 1,619 evaluable videos. HunyuanVideo-1.5 is excluded. \\
\addlinespace[7pt]
Track cross-check & Common 398 pairs measurable in all three Wan2.2-Turbo conditions; main BMR uses all 515. \\
\addlinespace[7pt]
Wan2.2-Turbo overlap and shape controls & 103 prompts $\times$ 5 seeds per support, paired with the main Wan2.2-Turbo Partial; generated with the control runner, which reproduces main Partial videos bit for bit. \\
\addlinespace[7pt]
Frame Guidance (CogVideoX-5B-I2V) & 103 prompts $\times$ 5 seeds per condition; Endpoint-free, FG-Full, and FG-$M_0$ are compared only with one another. \\
\addlinespace[7pt]
\bottomrule
\end{tabularx}
\end{table}

\FloatBarrier
\section{Paired Uncertainty and Additional Results}
\label{app:statistics}

\subsection{Resampling Protocol}

For a metric $y$, let $y_{isc}$ denote its value for prompt $i$, seed $s$, and condition $c$. The paired difference between target $a$ and reference $b$ is
\begin{equation}
\widehat{\Delta}_{a-b}=\frac{1}{103}\sum_{i=1}^{103}d_i,
\qquad
d_i=\frac15\sum_{s=1}^{5}(y_{isa}-y_{isb}).
\end{equation}
We resample 103 prompts with replacement, retaining each prompt's five seeds and paired conditions. We use 100,000 draws with NumPy PCG64 seed 20260919 and take the 2.5th/97.5th percentiles. TCR and temporal metrics share the same draws, which also give subgroup estimates: with $c_i^{(b)}$ the number of times prompt $i$ appears in draw $b$, the estimate for a prompt subgroup $g$, such as an edit type, is $\sum_{i\in g}c_i^{(b)}d_i/\sum_{i\in g}c_i^{(b)}$. TCR/DD differences are in pp, HC uses its displayed scale, and BMR its original scale.

Intervals are pointwise, unadjusted, and conditional on these five seeds. A CI spanning zero does not establish equivalence or non-inferiority. Because TCR differences take discrete values, a percentile bound can fall exactly at zero (e.g., $[0.00, +11.46]$ in Table~\ref{tab:main_ci}); such an interval counts as including zero. We compute differences and ratios from unrounded per-video values.

\begin{table}[!tbp]
\centering
\setlength{\tabcolsep}{4pt}
\newcommand{\resultcell}[2][c]{\begin{tabular}[c]{@{}#1@{}}#2\end{tabular}}
\caption{Paired mean differences. Each bootstrap draw resamples prompts and carries all five seeds and conditions together. The direction is stated in every row.}
\label{tab:main_ci}
\fitbox{%
\begin{tabular}{lcccc}
\toprule
 & \multicolumn{1}{c}{Transition completion} & \multicolumn{3}{c}{Temporal diagnostics} \\
\cmidrule(lr){2-2} \cmidrule(l){3-5}
Difference & $\Delta$TCR $\uparrow$ & $\Delta$HC $\downarrow$ & $\Delta$DD $\uparrow$ & $\Delta$BMR $\downarrow$ \\

\midrule
\resultcell[l]{Wan2.2-Turbo\\Full--Endpoint-free} & \resultcell{$+8.16$\\$[+1.75, +14.56]$} & \resultcell{$+1.90$\\$[+1.27, +2.59]$} & \resultcell{$-12.62$\\$[-17.48, -8.16]$} & \resultcell{$+0.239$\\$[+0.189, +0.289]$} \\
\addlinespace[7pt]

\rowcolor{ForestGreen!12}\resultcell[l]{Wan2.2-Turbo\\Partial--Full} & \resultcell{$+0.97$\\$[-3.50, +5.44]$} & \resultcell{$-0.72$\\$[-1.29, -0.17]$} & \resultcell{$+9.90$\\$[+5.63, +14.56]$} & \resultcell{$-0.213$\\$[-0.262, -0.165]$} \\
\addlinespace[7pt]

\resultcell[l]{Wan2.2-Turbo\\Partial--Endpoint-free} & \resultcell{$+9.13$\\$[+3.11, +15.34]$} & \resultcell{$+1.18$\\$[+0.72, +1.69]$} & \resultcell{$-2.72$\\$[-5.24, -0.39]$} & \resultcell{$+0.027$\\$[+0.004, +0.049]$} \\
\addlinespace[7pt]

\midrule
\resultcell[l]{SANA-Video\\Full--Endpoint-free} & \resultcell{$+12.23$\\$[+5.44, +19.22]$} & \resultcell{$+1.70$\\$[+0.67, +2.70]$} & \resultcell{$-9.90$\\$[-15.53, -4.27]$} & \resultcell{$+0.298$\\$[+0.254, +0.343]$} \\
\addlinespace[7pt]

\rowcolor{ForestGreen!12}\resultcell[l]{SANA-Video\\Partial--Full} & \resultcell{$-0.39$\\$[-6.41, +5.44]$} & \resultcell{$-0.97$\\$[-1.93, +0.01]$} & \resultcell{$+9.51$\\$[+4.08, +15.15]$} & \resultcell{$-0.285$\\$[-0.325, -0.246]$} \\
\addlinespace[7pt]

\resultcell[l]{SANA-Video\\Partial--Endpoint-free} & \resultcell{$+11.84$\\$[+6.99, +16.89]$} & \resultcell{$+0.73$\\$[+0.17, +1.33]$} & \resultcell{$-0.39$\\$[-4.85, +4.08]$} & \resultcell{$+0.013$\\$[-0.012, +0.038]$} \\
\addlinespace[7pt]

\midrule
\resultcell[l]{HunyuanVideo-1.5\\Full--Endpoint-free} & \resultcell{$+5.63$\\$[0.00, +11.46]$} & \resultcell{$+9.50$\\$[+8.53, +10.51]$} & \resultcell{$+1.17$\\$[-1.55, +3.88]$} & \resultcell{$+0.279$\\$[+0.233, +0.326]$} \\
\addlinespace[7pt]

\rowcolor{ForestGreen!12}\resultcell[l]{HunyuanVideo-1.5\\Partial--Full} & \resultcell{$-2.14$\\$[-7.57, +3.11]$} & \resultcell{$-7.27$\\$[-8.16, -6.40]$} & \resultcell{$-1.17$\\$[-3.50, +1.17]$} & \resultcell{$-0.252$\\$[-0.296, -0.208]$} \\
\addlinespace[7pt]

\resultcell[l]{HunyuanVideo-1.5\\Partial--Endpoint-free} & \resultcell{$+3.50$\\$[-0.58, +7.57]$} & \resultcell{$+2.23$\\$[+1.65, +2.87]$} & \resultcell{$+0.00$\\$[-1.75, +1.75]$} & \resultcell{$+0.028$\\$[+0.008, +0.048]$} \\
\addlinespace[7pt]

\bottomrule
\end{tabular}}
\end{table}

\subsection{Global-Softening and Spatial-Control Results}
\begin{table}[!tbp]
\centering
\setlength{\tabcolsep}{4pt}
\caption{Global softening versus Partial on Wan2.2-Turbo. Partial has higher DD and lower HC/BMR than each tested soft setting; all direct paired temporal intervals exclude zero (Table~\ref{tab:soft_ci}).}
\label{tab:soft}
\fitbox{%
\begin{tabular}{lrrrrr}
\toprule
 & \multicolumn{2}{c}{Transition completion} & \multicolumn{3}{c}{Temporal diagnostics} \\
\cmidrule(lr){2-3} \cmidrule(l){4-6}
Condition & TCR $\uparrow$ & TC-Score $\uparrow$ & HC $\downarrow$ & DD $\uparrow$ & BMR $\downarrow$ \\

\midrule
Full ($\alpha=1$) & 41.94 & 0.8352 & 6.47 & 66.8 & 0.599 \\

Soft ($\alpha=0.75$) & 42.33 & 0.8280 & 7.25 & 69.3 & 0.611 \\

Soft ($\alpha=0.5$) & 39.03 & 0.8155 & 7.62 & 69.9 & 0.630 \\

Soft ($\alpha=0.25$) & 39.22 & 0.7997 & 8.48 & 72.2 & 0.614 \\

Partial ($M_0$) & 42.91 & 0.8187 & 5.76 & 76.7 & 0.387 \\

\bottomrule
\end{tabular}}
\end{table}

\begin{table}[!tbp]
\centering
\setlength{\tabcolsep}{3pt}
\caption{Direct Partial minus Soft paired differences.}
\label{tab:soft_ci}
\fitbox{%
\begin{tabular}{@{}lcccc@{}}
\toprule
 & \multicolumn{1}{c}{Transition completion} & \multicolumn{3}{c}{Temporal diagnostics} \\
\cmidrule(lr){2-2} \cmidrule(l){3-5}
Soft $\alpha$ & $\Delta$TCR $\uparrow$ & $\Delta$HC $\downarrow$ & $\Delta$DD $\uparrow$ & $\Delta$BMR $\downarrow$ \\

\midrule
$0.75$ & $+0.58\;[-4.66, +5.63]$ & $-1.49\;[-2.23, -0.84]$ & $+7.38\;[+3.88, +11.07]$ & $-0.224\;[-0.272, -0.177]$ \\

$0.5$ & $+3.88\;[-1.75, +9.51]$ & $-1.86\;[-2.67, -1.16]$ & $+6.80\;[+2.91, +10.87]$ & $-0.243\;[-0.292, -0.195]$ \\

$0.25$ & $+3.69\;[-1.75, +9.13]$ & $-2.72\;[-3.60, -1.95]$ & $+4.47\;[+0.78, +8.35]$ & $-0.227\;[-0.273, -0.182]$ \\

\bottomrule
\end{tabular}}
\end{table}

\begin{table}[!tbp]
\centering
\setlength{\tabcolsep}{4pt}
\caption{Wan2.2-Turbo support-size and placement controls. Differences are paired with the main Partial.}
\label{tab:support}
\fitbox{%
\begin{tabular}{lrlrrr}
\toprule
 & \multicolumn{2}{c}{Transition completion} & \multicolumn{3}{c}{Temporal diagnostics} \\
\cmidrule(lr){2-3} \cmidrule(l){4-6}
Support & TCR $\uparrow$ & $\Delta$TCR $\uparrow$ [95\% CI] & HC $\downarrow$ & DD $\uparrow$ & BMR $\downarrow$ \\

\midrule
Partial ($M_0$) & 42.91 & -- & 5.76 & 76.7 & 0.387 \\

\midrule
$0.5A$ & 39.42 & $-3.50\;[-8.16, +0.97]$ & 5.23 & 77.1 & 0.370 \\

$0.75A$ & 42.72 & $-0.19\;[-4.08, +3.69]$ & 5.42 & 77.1 & 0.374 \\

$1.25A$ & 38.45 & $-4.47\;[-8.54, -0.39]$ & 5.86 & 76.3 & 0.405 \\

$1.5A$ & 41.94 & $-0.97\;[-4.85, +2.91]$ & 5.76 & 76.9 & 0.422 \\

\midrule
Shifted $M_0$ & 36.31 & $-6.60\;[-11.84, -1.55]$ & 5.31 & 77.3 & 0.428 \\

Random $A$ \#1 & 37.67 & $-5.24\;[-9.90, -0.78]$ & 5.72 & 75.3 & 0.414 \\

Random $A$ \#2 & 35.73 & $-7.18\;[-12.62, -1.75]$ & 5.80 & 75.0 & 0.418 \\

Random $A$ \#3 & 36.70 & $-6.21\;[-11.26, -1.17]$ & 5.79 & 75.3 & 0.404 \\

\bottomrule
\end{tabular}}
\end{table}

\begin{table}[!tbp]
\centering
\setlength{\tabcolsep}{4pt}
\caption{Support-control differences relative to the main Partial in Table~\ref{tab:support}. Near-zero bounds retain enough digits to preserve their sign.}
\label{tab:support_ci}
\fitbox{%
\begin{tabular}{lccc}
\toprule
 & \multicolumn{3}{c}{Temporal diagnostics} \\
\cmidrule(l){2-4}
Support & $\Delta$HC $\downarrow$ & $\Delta$DD $\uparrow$ & $\Delta$BMR $\downarrow$ \\

\midrule
$0.5A$ & $-0.53\;[-0.83, -0.25]$ & $+0.39\;[-1.36, +2.14]$ & $-0.017\;[-0.033, -0.001]$ \\

$0.75A$ & $-0.34\;[-0.68, -0.02]$ & $+0.39\;[-1.17, +1.94]$ & $-0.013\;[-0.026, +0.001]$ \\

$1.25A$ & $+0.10\;[-0.15, +0.36]$ & $-0.39\;[-2.33, +1.36]$ & $+0.019\;[+0.001, +0.036]$ \\

$1.5A$ & $+0.00\;[-0.28, +0.29]$ & $+0.19\;[-1.36, +1.94]$ & $+0.036\;[+0.017, +0.056]$ \\

\midrule
Shifted $M_0$ & $-0.44\;[-0.86, -0.03]$ & $+0.58\;[-1.94, +3.11]$ & $+0.041\;[+0.018, +0.066]$ \\

Random $A$ \#1 & $-0.04\;[-0.41, +0.33]$ & $-1.36\;[-3.50, +0.78]$ & $+0.028\;[+0.002, +0.054]$ \\

Random $A$ \#2 & $+0.05\;[-0.35, +0.45]$ & $-1.75\;[-3.88, +0.39]$ & $+0.031\;[+0.008, +0.056]$ \\

Random $A$ \#3 & $+0.03\;[-0.39, +0.45]$ & $-1.36\;[-3.69, +0.97]$ & $+0.017\;[-0.010, +0.045]$ \\

\bottomrule
\end{tabular}}
\end{table}

Tables~\ref{tab:main_ci}, \ref{tab:soft_ci}, and~\ref{tab:support_ci} report direct paired differences for the main, softening, and spatial comparisons.

\subsection{Placement Beyond Chance Coverage}
\label{app:placement}

\paragraph{Chance coverage.}
Random and shifted supports are placed without reference to $\Mzero$, yet they cover part of it by chance: random supports cover 9.0\% of $\Mzero$ on average (median 6\%), and shifted supports, despite IoU below 0.1, overlap $\Mzero$ in 62 of 103 prompts (8.0\% on average). When the draws are grouped by this coverage (Table~\ref{tab:coverage}), supports that cover little of $\Mzero$ lose the endpoint's completion gain, shifted supports with zero coverage fall to the Endpoint-free level, and supports that cover more of $\Mzero$ retain more of the gain. Coverage also varies with $\Mzero$'s size across prompts, so we control it directly.

\begin{table}[!tbp]
\centering
\setlength{\tabcolsep}{4pt}
\caption{Matched-area supports by the share of $M_0$ they cover by chance, $|S\cap M_0|/|M_0|$. Random draws (three per prompt, $309$ prompt--draw pairs) are split into quartiles of coverage; shifted supports into zero, partial, and higher overlap. $\Delta$ is relative to Partial; kept is the share of the endpoint's completion gain retained, $(\mathrm{TCR}_S-\mathrm{TCR}_{\rm EF})/(\mathrm{TCR}_{\rm Partial}-\mathrm{TCR}_{\rm EF})$.}
\label{tab:coverage}
\fitbox{%
\begin{tabular}{@{}lrrrcr@{}}
\toprule
Support, coverage & Pairs (prompts) & Mean coverage & $M_0$ area & $\Delta$TCR $\uparrow$ [95\% CI] & Kept \\
\midrule
\multicolumn{6}{@{}l}{\textit{Random $A$ ($2\times2$ blocks)}} \\
\quad $0.0$--$2.2\%$ & 77 (46) & $0.5\%$ & $2.6\%$ & $-9.09^{*}\;[-16.62, -1.45]$ & $-0.59$ \\
\quad $2.2$--$6.4\%$ & 77 (52) & $4.2\%$ & $4.5\%$ & $-8.31^{*}\;[-15.70, -2.31]$ & $-0.07$ \\
\quad $6.4$--$11.9\%$ & 77 (51) & $8.9\%$ & $7.3\%$ & $-5.19\hphantom{{}^{*}}\;[-13.00, +2.47]$ & $+0.44$ \\
\quad $11.9$--$54.0\%$ & 78 (34) & $22.4\%$ & $20.9\%$ & $-2.31\hphantom{{}^{*}}\;[-10.86, +7.13]$ & $+0.83$ \\
\multicolumn{6}{@{}l}{\textit{Shifted $M_0$}} \\
\quad $0\%$ & -- (41) & -- & $4.1\%$ & $-12.68^{*}\;[-22.50, -3.75]$ & $-0.04$ \\
\quad $0$--$8\%$ & -- (27) & -- & $8.1\%$ & $-1.48\hphantom{{}^{*}}\;[-9.47, +6.67]$ & $+0.33$ \\
\quad $>8\%$ & -- (35) & -- & $15.1\%$ & $-3.43\hphantom{{}^{*}}\;[-11.25, +5.00]$ & $+0.68$ \\
\bottomrule
\end{tabular}}
\end{table}

\paragraph{Controlled supports.}
Every set has exactly $|\Mzero|$ cells on the $44\times80$ latent grid. An overlap set covers a fraction $o\in\{0,0.25,0.5,0.75\}$ of $\Mzero$ with one contiguous sub-blob grown inside $\Mzero$ from a random $\Mzero$ cell and places the remaining cells as one contiguous blob outside $\Mzero$; $o=1$ is Partial. Three shape sets disregard $\Mzero$: one contiguous blob grown from a random cell, and scattered $4\times4$ or $8\times8$ blocks on the 4- and 8-cell lattices, versus $2\times2$ blocks for the main random supports. Seeds are derived from the prompt and set. All sets use the main Wan2.2 sampler through the control runner, which reproduces the main Partial videos bit for bit, and are paired with the main Partial.

\begin{figure}[!tbp]
\centering
\includegraphics[width=\linewidth]{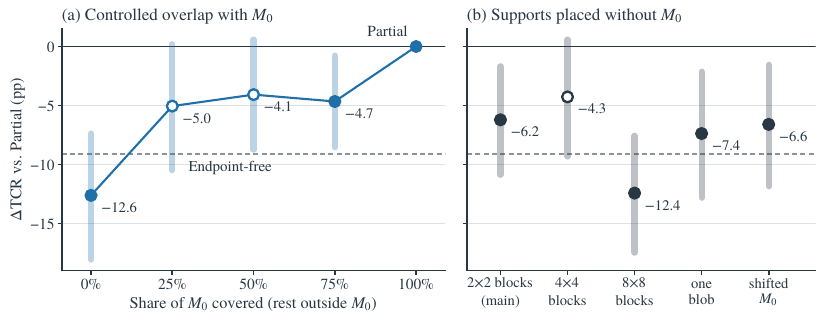}
\caption{\textbf{Completion requires covering the edit-relevant support.} Wan2.2-Turbo, paired $\Delta$TCR relative to Partial with 95\% CIs over 103 prompts and five seeds; open markers are not significant. (a) Supports with exactly $|\Mzero|$ cells covering 0--75\% of $\Mzero$, the rest placed outside it; 100\% is Partial. (b) Supports placed without reference to $\Mzero$, which cover 8--10\% of it by chance. The dashed line marks Endpoint-free.}
\label{fig:placement}
\end{figure}

\paragraph{Results.}
A support that misses $\Mzero$ entirely lowers TCR by 12.62~pp, below Endpoint-free, matching the zero-coverage shifted supports ($-12.68$~pp). Covering 25--75\% of $\Mzero$ retains 45--55\% of the endpoint's completion gain, and only the full region recovers the rest: even 75\% coverage stays significantly below Partial ($-4.66$~pp). All four random shapes lower TCR, three significantly, and all raise BMR significantly (Table~\ref{tab:overlap}; Fig.~\ref{fig:placement}). HC rises steadily with coverage, from 4.95 at zero overlap to 5.76 for Partial, consistent with HC arising where endpoint content is imposed; DD stays comparable throughout.

\begin{table}[!tbp]
\centering
\setlength{\tabcolsep}{3.5pt}
\caption{Wan2.2-Turbo supports with exactly $|M_0|$ cells: controlled overlap with $M_0$ and support shape. Differences are paired with the main Partial (103 prompts $\times$ five seeds); kept is defined as in Table~\ref{tab:coverage}. HC is multiplied by 100.}
\label{tab:overlap}
\fitbox{%
\begin{tabular}{@{}lrrcrccc@{}}
\toprule
Support & Coverage & TCR $\uparrow$ & $\Delta$TCR $\uparrow$ [95\% CI] & Kept & $\Delta$HC $\downarrow$ & $\Delta$DD $\uparrow$ & $\Delta$BMR $\downarrow$ \\
\midrule
Partial ($M_0$) & 100\% & 42.91 & -- & -- & -- & -- & -- \\
\midrule
Overlap 75\% & 75\% & 38.25 & $-4.66^{*}\;[-8.54, -0.78]$ & $+0.49$ & $-0.23\hphantom{{}^{*}}$ & $+0.39\hphantom{{}^{*}}$ & $+0.013\hphantom{{}^{*}}$ \\
Overlap 50\% & 50\% & 38.83 & $-4.08\hphantom{{}^{*}}\;[-8.74, +0.58]$ & $+0.55$ & $-0.44^{*}$ & $-1.17\hphantom{{}^{*}}$ & $+0.015\hphantom{{}^{*}}$ \\
Overlap 25\% & 25\% & 37.86 & $-5.05\hphantom{{}^{*}}\;[-10.49, +0.19]$ & $+0.45$ & $-0.65^{*}$ & $-0.97\hphantom{{}^{*}}$ & $+0.021\hphantom{{}^{*}}$ \\
Overlap 0\% & 0\% & 30.29 & $-12.62^{*}\;[-18.06, -7.38]$ & $-0.38$ & $-0.81^{*}$ & $-0.39\hphantom{{}^{*}}$ & $+0.025^{*}$ \\
\midrule
Random $2\times2$ (main, 3 draws) & 9.0\% & 36.70 & $-6.21^{*}\;[-10.87, -1.68]$ & $+0.32$ & $+0.01\hphantom{{}^{*}}$ & $-1.49\hphantom{{}^{*}}$ & $+0.025^{*}$ \\
Random $4\times4$ blocks & 9.2\% & 38.64 & $-4.27\hphantom{{}^{*}}\;[-9.32, +0.58]$ & $+0.53$ & $-0.51^{*}$ & $-0.97\hphantom{{}^{*}}$ & $+0.023^{*}$ \\
Random $8\times8$ blocks & 9.2\% & 30.49 & $-12.43^{*}\;[-17.48, -7.57]$ & $-0.36$ & $-0.77^{*}$ & $-1.36\hphantom{{}^{*}}$ & $+0.033^{*}$ \\
One blob, random place & 10.3\% & 35.53 & $-7.38^{*}\;[-12.82, -2.14]$ & $+0.19$ & $-0.90^{*}$ & $-0.19\hphantom{{}^{*}}$ & $+0.025^{*}$ \\
Shifted $M_0$ (main) & 8.0\% & 36.31 & $-6.60^{*}\;[-11.84, -1.55]$ & $+0.28$ & $-0.44^{*}$ & $+0.58\hphantom{{}^{*}}$ & $+0.041^{*}$ \\
\midrule
Endpoint-free & -- & 33.79 & -- & -- & -- & -- & -- \\
\bottomrule
\end{tabular}}
\end{table}

\paragraph{Static versus moving edit targets.}
Placement matters most when the edited object stays in place. We split prompts at the median optical flow inside $\Mzero$, with the split fixed before analysis. Among the 51 prompts below the median, shifted and random supports lower TCR by 10.20 and 9.67~pp (both significant), against 3.08 and 2.82~pp among the other 52. By contrast, Partial's BMR reduction relative to Full is the same in both groups ($-0.214$ and $-0.211$). Across edit types, the endpoint's completion gain comes from attribute binding and background shifts ($+16.0$~pp each), where random and shifted supports retain only 50--58\% of it.

\subsection{Additional Endpoint Comparisons}
\label{app:additional}

Additional endpoint-free versus Full comparisons use MfM, OmniWeaving, and Wan2.2 Base \citep{mfm,omniweaving,wan}.

\begin{table}[!tbp]
\centering
\setlength{\tabcolsep}{4pt}
\caption{Full--Endpoint-free completion differences. All rows use FLUX.2 endpoints except the ChronoEdit and Qwen-Image-Edit rows.}
\label{tab:endpoint_tcr}
\fitbox{%
\begin{tabular}{lrrl}
\toprule
 & \multicolumn{3}{c}{Transition completion} \\
\cmidrule(l){2-4}
Backbone & \shortstack{Endpoint-free\\TCR $\uparrow$} & Full TCR $\uparrow$ & Full--Endpoint-free [95\% CI] \\

\midrule
Wan2.2-Turbo & 33.79 & 41.94 & $+8.16\;[+1.75, +14.56]$ \\

Wan2.2-Turbo (ChronoEdit) & 33.79 & 40.19 & $+6.41\;[-0.19, +13.01]$ \\
Wan2.2-Turbo (Qwen-Image-Edit) & 33.79 & 35.15 & $+1.36\;[-5.63, +8.54]$ \\

SANA-Video & 23.50 & 35.73 & $+12.23\;[+5.44, +19.22]$ \\

HunyuanVideo-1.5 & 40.97 & 46.60 & $+5.63\;[0.00, +11.46]$ \\

MfM & 32.62 & 43.88 & $+11.26\;[+3.88, +18.64]$ \\

OmniWeaving & 39.42 & 43.30 & $+3.88\;[-2.52, +10.49]$ \\

Wan2.2 Base & 31.07 & 45.24 & $+14.17\;[+8.16, +20.39]$ \\

\bottomrule
\end{tabular}}
\end{table}

\begin{table}[!tbp]
\centering
\setlength{\tabcolsep}{4pt}
\newcommand{\resultcell}[2][c]{\begin{tabular}[c]{@{}#1@{}}#2\end{tabular}}
\caption{Wan2.2-Turbo completion by edit type: TCR means and paired differences. Partial uses the main fixed-support cohort.}
\label{tab:subgroups}
\fitbox{%
\begin{tabular}{lrrrccc}
\toprule
 & \multicolumn{6}{c}{Transition completion} \\
\cmidrule(l){2-7}
Edit type & \resultcell{Endpoint-\\free} & Full & Partial & \resultcell{Full--\\Endpoint-free} & \resultcell{Partial--\\Endpoint-free} & Partial--Full \\

\midrule
Attribute binding & 31.60 & 48.40 & 47.60 & \resultcell{$+16.80$\\$[+7.27, +26.79]$} & \resultcell{$+16.00$\\$[+5.56, +26.67]$} & \resultcell{$-0.80$\\$[-6.55, +4.58]$} \\
\addlinespace[7pt]

Background shifts & 38.67 & 46.67 & 54.67 & \resultcell{$+8.00$\\$[-5.33, +23.33]$} & \resultcell{$+16.00$\\$[+3.08, +30.91]$} & \resultcell{$+8.00$\\$[0.00, +20.00]$} \\
\addlinespace[7pt]

Object binding & 34.74 & 31.58 & 32.11 & \resultcell{$-3.16$\\$[-13.02, +6.45]$} & \resultcell{$-2.63$\\$[-9.00, +3.89]$} & \resultcell{$+0.53$\\$[-8.12, +9.41]$} \\
\addlinespace[7pt]

\midrule
Overall & 33.79 & 41.94 & 42.91 & \resultcell{$+8.16$\\$[+1.75, +14.56]$} & \resultcell{$+9.13$\\$[+3.11, +15.34]$} & \resultcell{$+0.97$\\$[-3.50, +5.44]$} \\
\addlinespace[7pt]

\bottomrule
\end{tabular}}
\end{table}

Full--Endpoint-free TCR gains are significant for Wan2.2, SANA, MfM, and Wan2.2 Base, and comparable for Wan2.2 with ChronoEdit or Qwen-Image-Edit endpoints, Hunyuan, and OmniWeaving (Table~\ref{tab:endpoint_tcr}).

In Wan2.2's edit-type results, Partial--Endpoint-free intervals are positive for Attribute binding and Background shifts, while the interval for Object binding spans zero. All subgroup Partial--Full TCR intervals include zero (Table~\ref{tab:subgroups}), while Partial's BMR reduction relative to Full is significant in every edit type on all three backbones (Table~\ref{tab:subgroups_temporal}).

\begin{table}[!tbp]
\centering
\setlength{\tabcolsep}{3.5pt}
\caption{Partial--Full differences by edit type on all three backbones (50 attribute-binding, 38 object-binding, and 15 background-shift prompts; five seeds each). Subgroup intervals follow Appendix~\ref{app:statistics}.}
\label{tab:subgroups_temporal}
\fitbox{%
\begin{tabular}{@{}llcccc@{}}
\toprule
 & & \multicolumn{1}{c}{Transition completion} & \multicolumn{3}{c}{Temporal diagnostics} \\
\cmidrule(lr){3-3} \cmidrule(l){4-6}
Backbone & Edit type & $\Delta$TCR $\uparrow$ & $\Delta$HC $\downarrow$ & $\Delta$DD $\uparrow$ & $\Delta$BMR $\downarrow$ \\
\midrule
Wan2.2-Turbo & Attribute binding & $-0.80\hphantom{{}^{*}}\;[-6.55, +4.58]$ & $-0.68\hphantom{{}^{*}}\;[-1.57, +0.16]$ & $+9.20^{*}\;[+3.50, +15.56]$ & $-0.201^{*}\;[-0.256, -0.147]$ \\
 & Object binding & $+0.53\hphantom{{}^{*}}\;[-8.12, +9.41]$ & $-0.50\hphantom{{}^{*}}\;[-1.30, +0.24]$ & $+10.53^{*}\;[+2.94, +19.38]$ & $-0.210^{*}\;[-0.308, -0.115]$ \\
 & Background shifts & $+8.00\hphantom{{}^{*}}\;[0.00, +20.00]$ & $-1.38\hphantom{{}^{*}}\;[-3.18, +0.22]$ & $+10.67\hphantom{{}^{*}}\;[0.00, +23.64]$ & $-0.258^{*}\;[-0.403, -0.121]$ \\
\midrule
SANA-Video & Attribute binding & $-6.80\hphantom{{}^{*}}\;[-16.25, +2.46]$ & $-0.07\hphantom{{}^{*}}\;[-1.39, +1.18]$ & $+12.00^{*}\;[+5.42, +18.80]$ & $-0.309^{*}\;[-0.369, -0.249]$ \\
 & Object binding & $+4.21\hphantom{{}^{*}}\;[-5.29, +13.55]$ & $-2.38^{*}\;[-3.78, -1.00]$ & $+4.74\hphantom{{}^{*}}\;[-5.50, +15.56]$ & $-0.293^{*}\;[-0.356, -0.231]$ \\
 & Background shifts & $+9.33\hphantom{{}^{*}}\;[0.00, +20.00]$ & $-0.37\hphantom{{}^{*}}\;[-3.59, +3.66]$ & $+13.33\hphantom{{}^{*}}\;[-3.64, +30.00]$ & $-0.184^{*}\;[-0.260, -0.107]$ \\
\midrule
HunyuanVideo-1.5 & Attribute binding & $-4.00\hphantom{{}^{*}}\;[-11.30, +3.11]$ & $-6.82^{*}\;[-8.29, -5.39]$ & $0.00\hphantom{{}^{*}}\;[-3.64, +3.75]$ & $-0.316^{*}\;[-0.371, -0.259]$ \\
 & Object binding & $0.00\hphantom{{}^{*}}\;[-10.67, +10.45]$ & $-7.67^{*}\;[-8.98, -6.44]$ & $-2.11\hphantom{{}^{*}}\;[-5.00, +0.62]$ & $-0.153^{*}\;[-0.206, -0.100]$ \\
 & Background shifts & $-1.33\hphantom{{}^{*}}\;[-10.00, +7.69]$ & $-7.75^{*}\;[-9.79, -5.93]$ & $-2.67\hphantom{{}^{*}}\;[-10.67, +4.44]$ & $-0.289^{*}\;[-0.465, -0.123]$ \\
\bottomrule
\end{tabular}}
\end{table}

\begin{table}[!tbp]
\centering
\setlength{\tabcolsep}{4pt}
\caption{Full--Endpoint-free temporal differences by endpoint-conditioning mechanism, with Endpoint-free levels (EF). Paired prompt bootstrap as in Table~\ref{tab:main_ci}; comparisons are paired within each model. Rows without an editor use FLUX.2 endpoints.}
\label{tab:learned}
\fitbox{%
\begin{tabular}{@{}lrcrcrc@{}}
\toprule
 & \multicolumn{2}{c}{HC $\downarrow$} & \multicolumn{2}{c}{DD $\uparrow$ (\%)} & \multicolumn{2}{c}{BMR $\downarrow$} \\
\cmidrule(lr){2-3}\cmidrule(lr){4-5}\cmidrule(l){6-7}
Model & EF & $\Delta$ [95\% CI] & EF & $\Delta$ [95\% CI] & EF & $\Delta$ [95\% CI] \\
\midrule
\multicolumn{7}{@{}l}{\textit{Training-free terminal projection}} \\
Wan2.2-Turbo & 4.57 & $+1.90\;[+1.27, +2.59]$ & 79.4 & $-12.62\;[-17.48, -8.16]$ & 0.360 & $+0.239\;[+0.189, +0.289]$ \\
SANA-Video & 8.07 & $+1.70\;[+0.67, +2.70]$ & 42.7 & $-9.90\;[-15.53, -4.27]$ & 0.122 & $+0.298\;[+0.254, +0.343]$ \\
HunyuanVideo-1.5 & 3.47 & $+9.50\;[+8.53, +10.51]$ & 51.5 & $+1.17\;[-1.55, +3.88]$ & 0.305 & $+0.279\;[+0.233, +0.326]$ \\
Wan2.2 Base & 5.71 & $+1.47\;[+0.76, +2.24]$ & 60.0 & $-2.14\;[-6.99, +2.72]$ & 0.415 & $+0.082\;[+0.041, +0.123]$ \\
Wan2.2-Turbo (Qwen-Image-Edit) & 4.57 & $+2.54\;[+1.80, +3.35]$ & 79.4 & $-15.34\;[-20.19, -10.68]$ & 0.360 & $+0.256\;[+0.207, +0.307]$ \\
Wan2.2-Turbo (ChronoEdit) & 4.57 & $+1.55\;[+0.93, +2.18]$ & 79.4 & $-11.46\;[-16.89, -6.02]$ & 0.360 & $+0.198\;[+0.148, +0.249]$ \\
\midrule
\multicolumn{7}{@{}l}{\textit{Learned endpoint conditioning}} \\
MfM & 3.61 & $+0.52\;[-0.02, +1.12]$ & 48.3 & $-20.19\;[-27.57, -13.20]$ & 0.368 & $+0.019\;[-0.025, +0.063]$ \\
OmniWeaving & 6.06 & $-0.41\;[-0.90, +0.08]$ & 54.6 & $-17.67\;[-24.27, -11.07]$ & 0.289 & $+0.198\;[+0.149, +0.245]$ \\
\bottomrule
\end{tabular}}
\end{table}

\paragraph{Temporal artifacts across conditioning mechanisms.}
Table~\ref{tab:learned} extends the Full--Endpoint-free comparison to temporal diagnostics. Wan2.2 (with FLUX.2, Qwen-Image-Edit, or ChronoEdit endpoints), SANA, Hunyuan, and Wan2.2 Base impose the edited endpoint through training-free terminal projection, whereas MfM and OmniWeaving learn endpoint conditioning during training. Every Full setting incurs at least one significant temporal artifact, but its form differs. Training-free projection raises HC and BMR in all six settings. Learned conditioning leaves HC comparable but lowers DD by 20.19~pp for MfM and 17.67~pp for OmniWeaving; OmniWeaving also raises BMR by 0.198. Learned conditioning therefore does not remove the temporal artifacts of full-frame endpoints; it shifts them toward motion suppression. Comparisons are paired within each model, whose frame counts and resolutions differ.

\subsection{Frame Guidance with Edit-Relevant Support}
\label{app:fg}

Frame Guidance steers a video model toward target frames by backpropagating a frame loss through the VAE decoder during sampling \citep{frameguidance}. We run its official keyframe configuration on its primary backbone, CogVideoX-5B-I2V \citep{cogvideox}: 49 frames at $480\times720$ and 8 FPS, 50 steps, and CFG 6, with guidance on the last frame only. Inputs are the same 103 prompts and five seeds (22001--22005); source images and FLUX.2 endpoints are center-cropped to 3:2, and $\Mzero$ is cropped accordingly. FG-Full guides the whole last frame toward the endpoint, FG-$\Mzero$ weights the same loss by the cropped $\Mzero$ through the official \texttt{mask\_video} option, and Endpoint-free runs the same sampler without guidance. BMR uses the cropped $\Mzero$.

Table~\ref{tab:fg} gives the paired intervals behind Section~\ref{sec:fg}. Relative to Endpoint-free, FG-$\Mzero$ keeps a TCR gain of $+10.10$~pp with comparable DD ($-1.94$~pp). FG-Full's BMR increase is small ($+0.045$, $[-0.001, +0.090]$), leaving no reversal for the restriction to remove. Of the 108 FG-$\Mzero$ videos with HC above 10, 39 have their minimum-similarity pair in the first second, against 15 of 75 for FG-Full; these early events are changes to unrelated scenes rather than jumps toward the endpoint.

These three conditions were judged and measured on A100 GPUs, with the same judge model and revision served through a vLLM build for CUDA 12.9 and the same metric code. On the 515 Wan2.2 Full videos, this setup reproduces the original verdicts for 96.7\% of videos, with a TCR difference of $-0.58$~pp $[-2.14, +0.78]$, within the variation of re-judging on the original machine (95.7\%; $-0.78$~pp $[-2.52, +0.97]$). The three Frame Guidance conditions are compared only with one another.

\begin{table}[!tbp]
\centering
\setlength{\tabcolsep}{3.5pt}
\providecommand{\resultcell}[2][c]{\begin{tabular}[c]{@{}#1@{}}#2\end{tabular}}
\caption{Frame Guidance on CogVideoX-5B-I2V (103 prompts $\times$ five seeds). FG-$M_0$ weights the same frame loss by the cropped $M_0$. Minutes per video on one A6000. HC is multiplied by 100.}
\label{tab:fg}
\fitbox{%
\begin{tabular}{@{}lccccr@{}}
\toprule
 & \multicolumn{1}{c}{Transition completion} & \multicolumn{3}{c}{Temporal diagnostics} & \\
\cmidrule(lr){2-2} \cmidrule(lr){3-5}
Condition & TCR $\uparrow$ & HC $\downarrow$ & DD $\uparrow$ & BMR $\downarrow$ & Min/video \\
\midrule
Endpoint-free & 27.18 & 5.44 & 44.7 & 0.388 & 7 \\
FG-Full (whole last frame) & 36.89 & 5.40 & 30.5 & 0.434 & 57 \\
FG-$M_0$ (last frame, $M_0$ only) & 37.28 & 6.90 & 42.7 & 0.427 & 57 \\
\midrule
FG-Full $-$ Endpoint-free & \resultcell{$+9.71^{*}$\\$[+3.69, +15.92]$} & \resultcell{$-0.04$\\$[-0.65, +0.59]$} & \resultcell{$-14.17^{*}$\\$[-20.19, -8.16]$} & \resultcell{$+0.045$\\$[-0.001, +0.090]$} & \\
\addlinespace[3pt]
FG-$M_0$ $-$ FG-Full & \resultcell{$+0.39$\\$[-3.69, +4.47]$} & \resultcell{$+1.50^{*}$\\$[+0.89, +2.14]$} & \resultcell{$+12.23^{*}$\\$[+7.18, +17.48]$} & \resultcell{$-0.006$\\$[-0.043, +0.032]$} & \\
\addlinespace[3pt]
FG-$M_0$ $-$ Endpoint-free & \resultcell{$+10.10^{*}$\\$[+4.47, +15.92]$} & \resultcell{$+1.46^{*}$\\$[+0.66, +2.30]$} & \resultcell{$-1.94$\\$[-6.99, +3.11]$} & \resultcell{$+0.039$\\$[-0.001, +0.079]$} & \\
\addlinespace[3pt]
\bottomrule
\end{tabular}}
\end{table}

\FloatBarrier
\section{Donor-Context Intervention and Score Decomposition}
\label{app:context}

\subsection{Construction, Pairing, and Scope}

The conditions separate terminal content from projection support:
\begin{table}[h]
\centering
\caption{Donor-context construction. O/D/$D_{\rm cross}$/N all retain Full support. Content inside $M_0$ always comes from the original edited endpoint.}
\label{tab:context_construction}
\begin{tabularx}{\linewidth}{@{}p{0.28\linewidth}XXl@{}}
\toprule
Condition & Inside $M_0$ & Outside $M_0$ & Support \\
\midrule
O: Original Full & Original latent & Original latent & Full \\
D: Donor-context Full & Original latent & Re-encoded donor final RGB & Full \\
$D_{\rm cross}$: Cross-seed donor & Original latent & Re-encoded final RGB of another seed's donor & Full \\
N: Native-latent Full & Original latent & Native donor tail latent & Full \\
P: Partial donor & Original latent & Video prediction retained & $M_0$ \\
\bottomrule
\end{tabularx}
\end{table}

\paragraph{Generation and pairing.}
On Hunyuan, we use 480p, eight steps, and 121 frames at 24 FPS, with the native scheduler and noise-matched projection. Each of the 103 prompts uses seeds 22001--22005. Within each pair, source, prompt, edited endpoint, transformed $M_0$, and initial/endpoint noise are fixed.

P is a Partial rollout. We take its final RGB frame before compression and its native terminal latent. We encode the original endpoint and donor RGB through the same five-frame repetition procedure and select the second temporal slot. D uses Eq.~\eqref{eq:future_context}; N uses the native donor slot instead. Both retain the original endpoint inside $M_0$.

O reuses the main Full videos; P, D, and N each contain 515 new videos. The donor P cohort is separate from main Partial (Table~\ref{tab:populations}).

On Wan2.2, all four conditions are new 515-video runs with the main four-step sampler, 121 frames at 24 FPS, and one CUDA generator per prompt--seed that drives the initial noise and per-step re-noising of every condition. P is Partial with the main settings; we save its final RGB frame before compression and its native terminal latent. The endpoint and donor RGB use the same tail-slot encoding. $\Mzero$ enters as the pooled weight map of Wan2.2 Partial ($2\times2$ average pooling to the $22\times40$ token grid, repeated to the $44\times80$ latent grid), so it selects content, while O, D, and N use Full support ($w=1$).

Both backbones use the same BMR measurement mask for all conditions and all 515 pairs, without DD filtering. Confidence intervals follow Appendix~\ref{app:statistics}.

\paragraph{Interpretation.}
\label{app:context_scope}
O$\rightarrow$D changes non-edit terminal content under fixed Full support. The donor is generated, not ground-truth future context; position/visibility, generated-image properties, and trajectory compatibility are not separately manipulated. D$\rightarrow$N tests terminal representation. Partial itself ignores endpoint content outside $M_0$, so replacing that content does not change its projection.

\begin{table}[!tbp]
\centering
\setlength{\tabcolsep}{5pt}
\caption{Donor-context condition means. Within each backbone, O/D/$D_{\rm cross}$/N use Full support; P is the Partial donor cohort. All conditions share the BMR measurement mask.}
\label{tab:context_means}
\fitbox{%
\begin{tabular}{lrrrrr}
\toprule
 & \multicolumn{2}{c}{Transition completion} & \multicolumn{3}{c}{Temporal diagnostics} \\
\cmidrule(lr){2-3} \cmidrule(l){4-6}
Condition & TCR $\uparrow$ & TC-Score $\uparrow$ & HC $\downarrow$ & DD $\uparrow$ & BMR $\downarrow$ \\
\midrule
\multicolumn{6}{@{}l}{\textit{HunyuanVideo-1.5}} \\
O --- Original Full & 46.60 & 0.859 & 12.96 & 52.62 & 0.585 \\
D --- Donor-context Full & 47.57 & 0.873 & 8.10 & 52.82 & 0.358 \\
$D_{\rm cross}$ --- Cross-seed donor & 46.80 & 0.869 & 11.27 & 52.04 & 0.480 \\
N --- Native-latent Full & 45.63 & 0.879 & 7.16 & 52.23 & 0.354 \\
P --- Partial & 44.66 & 0.882 & 5.69 & 51.46 & 0.333 \\
\midrule
\multicolumn{6}{@{}l}{\textit{Wan2.2-Turbo}} \\
O --- Original Full & 43.30 & 0.834 & 6.49 & 65.44 & 0.600 \\
D --- Donor-context Full & 43.11 & 0.821 & 6.12 & 75.34 & 0.362 \\
N --- Native-latent Full & 42.72 & 0.820 & 6.26 & 74.95 & 0.382 \\
P --- Partial & 42.72 & 0.819 & 5.76 & 76.70 & 0.387 \\
\bottomrule
\end{tabular}}
\end{table}

\begin{table}[!tbp]
\centering
\setlength{\tabcolsep}{3.5pt}
\providecommand{\resultcell}[2][c]{\begin{tabular}[c]{@{}#1@{}}#2\end{tabular}}
\caption{Paired donor-context differences. D--O and $D_{\rm cross}$--O are the primary contrasts; the others are secondary. Appendix~\ref{app:statistics} specifies the uncertainty interpretation.}
\label{tab:context_ci}
\fitbox{%
\begin{tabular}{@{}lccccc@{}}
\toprule
 & \multicolumn{2}{c}{Transition completion} & \multicolumn{3}{c}{Temporal diagnostics} \\
\cmidrule(lr){2-3} \cmidrule(l){4-6}
Difference & $\Delta$TCR $\uparrow$ & $\Delta$TC-Score $\uparrow$ & $\Delta$HC $\downarrow$ & $\Delta$DD $\uparrow$ & $\Delta$BMR $\downarrow$ \\
\midrule
\multicolumn{6}{@{}l}{\textit{HunyuanVideo-1.5}} \\
D--O & \resultcell{$+0.97$\\$[-4.66, +6.60]$} & \resultcell{$+0.014$\\$[+0.003, +0.026]$} & \resultcell{$-4.86$\\$[-5.75, -3.99]$} & \resultcell{$+0.19$\\$[-1.55, +1.94]$} & \resultcell{$-0.227$\\$[-0.267, -0.186]$} \\
\addlinespace[4pt]
$D_{\rm cross}$--O & \resultcell{$+0.19$\\$[-5.63, +6.02]$} & \resultcell{$+0.010$\\$[-0.000, +0.021]$} & \resultcell{$-1.70$\\$[-2.47, -0.94]$} & \resultcell{$-0.58$\\$[-2.14, +0.78]$} & \resultcell{$-0.105$\\$[-0.141, -0.068]$} \\
\addlinespace[4pt]
$D_{\rm cross}$--D & \resultcell{$-0.78$\\$[-3.88, +2.33]$} & \resultcell{$-0.004$\\$[-0.009, +0.001]$} & \resultcell{$+3.16$\\$[+2.65, +3.71]$} & \resultcell{$-0.78$\\$[-2.14, +0.58]$} & \resultcell{$+0.122$\\$[+0.091, +0.154]$} \\
\addlinespace[4pt]
N--O & \resultcell{$-0.97$\\$[-6.60, +4.66]$} & \resultcell{$+0.020$\\$[+0.010, +0.030]$} & \resultcell{$-5.80$\\$[-6.68, -4.94]$} & \resultcell{$-0.39$\\$[-1.94, +1.17]$} & \resultcell{$-0.231$\\$[-0.272, -0.189]$} \\
\addlinespace[4pt]
N--D & \resultcell{$-1.94$\\$[-5.44, +1.36]$} & \resultcell{$+0.005$\\$[-0.000, +0.011]$} & \resultcell{$-0.94$\\$[-1.17, -0.71]$} & \resultcell{$-0.58$\\$[-1.55, +0.39]$} & \resultcell{$-0.004$\\$[-0.018, +0.009]$} \\
\addlinespace[4pt]
O--P & \resultcell{$+1.94$\\$[-3.50, +7.38]$} & \resultcell{$-0.023$\\$[-0.034, -0.012]$} & \resultcell{$+7.27$\\$[+6.39, +8.17]$} & \resultcell{$+1.17$\\$[-1.17, +3.50]$} & \resultcell{$+0.251$\\$[+0.208, +0.295]$} \\
\addlinespace[4pt]
D--P & \resultcell{$+2.91$\\$[0.00, +5.83]$} & \resultcell{$-0.008$\\$[-0.014, -0.003]$} & \resultcell{$+2.41$\\$[+2.07, +2.75]$} & \resultcell{$+1.36$\\$[-0.97, +3.50]$} & \resultcell{$+0.025$\\$[+0.004, +0.046]$} \\
\midrule
\multicolumn{6}{@{}l}{\textit{Wan2.2-Turbo}} \\
D--O & \resultcell{$-0.19$\\$[-5.05, +4.27]$} & \resultcell{$-0.013$\\$[-0.024, -0.002]$} & \resultcell{$-0.37$\\$[-0.95, +0.18]$} & \resultcell{$+9.90$\\$[+5.63, +14.56]$} & \resultcell{$-0.238$\\$[-0.283, -0.195]$} \\
\addlinespace[4pt]
N--O & \resultcell{$-0.58$\\$[-5.44, +4.08]$} & \resultcell{$-0.013$\\$[-0.026, -0.001]$} & \resultcell{$-0.24$\\$[-0.74, +0.24]$} & \resultcell{$+9.51$\\$[+4.85, +14.37]$} & \resultcell{$-0.218$\\$[-0.263, -0.174]$} \\
\addlinespace[4pt]
N--D & \resultcell{$-0.39$\\$[-2.91, +2.14]$} & \resultcell{$-0.000$\\$[-0.007, +0.006]$} & \resultcell{$+0.14$\\$[-0.14, +0.41]$} & \resultcell{$-0.39$\\$[-2.14, +1.36]$} & \resultcell{$+0.020$\\$[+0.001, +0.040]$} \\
\addlinespace[4pt]
O--P & \resultcell{$+0.58$\\$[-4.47, +5.83]$} & \resultcell{$+0.015$\\$[+0.003, +0.027]$} & \resultcell{$+0.74$\\$[+0.22, +1.29]$} & \resultcell{$-11.26$\\$[-16.31, -6.60]$} & \resultcell{$+0.213$\\$[+0.170, +0.258]$} \\
\addlinespace[4pt]
D--P & \resultcell{$+0.39$\\$[-3.11, +4.08]$} & \resultcell{$+0.002$\\$[-0.005, +0.009]$} & \resultcell{$+0.36$\\$[+0.05, +0.69]$} & \resultcell{$-1.36$\\$[-3.88, +1.17]$} & \resultcell{$-0.025$\\$[-0.049, -0.001]$} \\
\bottomrule
\end{tabular}}
\end{table}

\subsection{Cross-Seed Donors on Hunyuan}
\label{app:context_cross}
For each prompt, a cyclic shift $k\in\{1,2,3,4\}$ assigns donor seed index $(j+k)\bmod 5$ to recipient seed index $j$. Shifts are balanced across prompts (26/26/26/25) and were fixed before generation. The recipient keeps its source, prompt, seed, initial and endpoint noises, and transformed $\Mzero$; only the donor frame used outside $\Mzero$ changes. A pilot reproduced the existing Original Full videos and, with the donor seed set to the recipient seed, the same-seed D videos exactly. Cross-seed donors retain 35\% ($[22, 45]$) of the same-seed HC reduction and 46\% ($[33, 58]$) of the BMR reduction (Tables~\ref{tab:context_means} and~\ref{tab:context_ci}).

\subsection{Secondary Donor and Representation Contrasts}

On Hunyuan, donor-context Full has $+2.91\;[0.00, +5.83]$~pp TCR relative to P, with higher HC and BMR; the completion interval includes zero. Replacing the re-encoded donor with its native slot further reduces HC ($-0.94\;[-1.17, -0.71]$), while the BMR interval spans zero (Table~\ref{tab:context_ci}). On Wan2.2, the native terminal latent leaves TCR, HC, and DD comparable to the re-encoded donor and raises BMR ($+0.020\;[+0.001, +0.040]$). Relative to P, D has comparable TCR and DD, slightly lower BMR, and slightly higher HC (Table~\ref{tab:context_ci}).

\subsection{Descriptive Closure of the Full--Partial Gap}
\label{app:context_gap}

For an artifact metric $m$, the two summaries use different denominators:
\begin{equation}
\text{Relative reduction}=\frac{\bar m_O-\bar m_D}{\bar m_O},\qquad
R_m=\text{Gap closure}=\frac{\bar m_O-\bar m_D}{\bar m_O-\bar m_P}.
\end{equation}
On Hunyuan, O$\rightarrow$D reduces HC by 37.5\% and BMR by 38.8\% relative to O. Using donor P as reference, it closes 66.83\% and 90.16\% of the Full--Partial gaps. Residual D--P differences are $+2.41\;[+2.07, +2.75]$ for HC and $+0.025\;[+0.004, +0.046]$ for BMR. On Wan2.2, with the same-run P as reference, D closes 112\% ($[100, 125]$) of the BMR gap, reaching P's BMR; the HC closure is uncertain (51\%, $[-58, 93]$) because the Full--Partial HC gap is small. Gap closure summarizes these mean differences; it is not a causal mediation estimate.

\subsection{TC-Score Decomposition and Its Relation to HC}
\label{app:context_tc}

For $\mathrm{TC\text{-}Score}=\frac23V+\frac13C$, we bootstrap the VLM term $V$ and mapped CLIP term $C$ over the same paired prompts. Their weighted O$\rightarrow$D contributions are $+0.00298$ and $+0.01150$, totaling $+0.01448$ (Table~\ref{tab:context_tc}). The increase is mainly in CLIP consistency; the VLM interval includes zero. On Wan2.2, the weighted contributions are $-0.00258$ and $-0.01031$, totaling $-0.01289$: the decrease is mainly in CLIP consistency, the VLM interval includes zero, and it accompanies a 9.90~pp DD increase. TC-Score and HC both use adjacent-frame CLIP similarity, so their agreement is not independent evidence of completion.

\begin{table}[!tbp]
\centering
\setlength{\tabcolsep}{5pt}
\caption{O$\rightarrow$D decomposition of TC-Score on each backbone. Weighted VLM and CLIP terms sum to TC-Score.}
\label{tab:context_tc}
\fitbox{%
\begin{tabular}{lrrl}
\toprule
Component & O mean & D mean & $\Delta$ [95\% CI] \\
\midrule
\multicolumn{4}{@{}l}{\textit{HunyuanVideo-1.5}} \\
VLM & 0.86545 & 0.86991 & $+0.00447\;[-0.01200, +0.02020]$ \\
Mapped CLIP & 0.84545 & 0.87995 & $+0.03451\;[+0.02604, +0.04341]$ \\
$\frac23$ VLM & 0.57696 & 0.57994 & $+0.00298\;[-0.00800, +0.01347]$ \\
$\frac13$ mapped CLIP & 0.28182 & 0.29332 & $+0.01150\;[+0.00868, +0.01447]$ \\
TC-Score $\uparrow$ & 0.85878 & 0.87326 & $+0.01448\;[+0.00293, +0.02575]$ \\
\midrule
\multicolumn{4}{@{}l}{\textit{Wan2.2-Turbo}} \\
VLM & 0.84855 & 0.84468 & $-0.00387\;[-0.01737, +0.00941]$ \\
Mapped CLIP & 0.80368 & 0.77274 & $-0.03093\;[-0.04778, -0.01484]$ \\
$\frac23$ VLM & 0.56570 & 0.56312 & $-0.00258\;[-0.01158, +0.00627]$ \\
$\frac13$ mapped CLIP & 0.26789 & 0.25758 & $-0.01031\;[-0.01593, -0.00495]$ \\
TC-Score $\uparrow$ & 0.83359 & 0.82070 & $-0.01289\;[-0.02386, -0.00233]$ \\
\bottomrule
\end{tabular}}
\end{table}

\FloatBarrier
\section{Diagnostics and Cross-Checks}
\label{app:validation}

\subsection{Hunyuan Discontinuity Localization}
\label{app:hc_localization}

\begin{table}[!tbp]
\centering
\setlength{\tabcolsep}{4pt}
\caption{HunyuanVideo-1.5 Dynamic Degree after removing HC-neighbor pairs or the three largest-flow pairs. The original motion threshold and count requirement remain fixed. HC-neighbor removal drops two pairs at a sequence edge and three otherwise.}
\label{tab:dd_sensitivity}
\fitbox{%
\begin{tabular}{lrrr}
\toprule
Condition & Original DD $\uparrow$ & Remove $t_{\rm HC}\pm1$ & Remove top-3 flow \\

\midrule
Endpoint-free & 51.5 & 48.3 & 46.0 \\

Full & 52.6 & 50.5 & 47.0 \\

Partial & 51.5 & 49.9 & 46.6 \\

\bottomrule
\end{tabular}}
\end{table}

\begin{table}[!tbp]
\centering
\setlength{\tabcolsep}{4pt}
\caption{HunyuanVideo-1.5 HC localization and RAFT magnitude. $t_{\rm HC}$ is each video's minimum-similarity pair. The last two columns average RAFT top-5\% magnitudes at that pair and at the remaining pairs; they do not use a fixed pair 38. The second column reports the fraction with $t_{\rm HC}=38$.}
\label{tab:hc_localization}
\fitbox{%
\begin{tabular}{lrrr}
\toprule
Condition & $t_{\rm HC}=38$ & $m(t_{\rm HC})$ & Other pairs \\

\midrule
Endpoint-free & 1.7\% & 19.5 px & 11.3 px \\

Full & 95.9\% & 88.9 px & 11.4 px \\

Partial & 51.8\% & 18.3 px & 11.4 px \\

\bottomrule
\end{tabular}}
\end{table}

Let $t_{\rm HC}$ be each video's minimum-similarity adjacent-frame pair on the 8-FPS grid. We remove either $t_{\rm HC}$ and its neighbors or the three largest-flow pairs while keeping DD's threshold and qualifying-pair count fixed (Table~\ref{tab:dd_sensitivity}). Both removals reduce DD similarly across Endpoint-free, Full, and Partial, so Full's maintained DD is not explained by a few extreme motion spikes. Pair 38 connects native frames 114 and 117, the onset of Hunyuan's terminal latent slot, and is $t_{\rm HC}$ in 95.9\% of Full videos, versus 1.7\% for Endpoint-free and 51.8\% for Partial. At each video's own $t_{\rm HC}$, mean RAFT top-5\% magnitude is 88.9 px for Full, compared with 19.5 and 18.3 px for Endpoint-free and Partial; all remaining pairs average 11.3--11.4 px (Table~\ref{tab:hc_localization}).

\FloatBarrier
\subsection{Blinded Model-Judge Agreement for HC}
\label{app:judge}

\paragraph{Protocol.}
We use Qwen3.6-35B-A3B with temperature 0, top-$p$ 1, thinking disabled, and a 1,536-token response limit. We evaluate 90 prompts outside the 13-prompt development subset, seeds 22001 and 22002, and nine conditions: Wan2.2 Endpoint-free, Full, Partial, adaptive release, and random release; MfM Full; OmniWeaving Full; and SANA Full and Partial. Hunyuan is excluded. The judge sees all native frames (81--121), resized on a 32-pixel grid within a $768\times432$ per-frame pixel budget. Method names, prompts, endpoints, masks, and metric values are hidden; no DD filter is used.

We ask whether any interval changes abruptly without a continuous transition. The judge provides visible evidence and event intervals before a PRESENT/ABSENT/UNCERTAIN label. Continuous fast motion and smooth fades are excluded, and first/last-frame differences alone are insufficient.

\paragraph{Results.}
Of 1,620 inputs, 1,619 receive valid judgments, 68 PRESENT and 1,551 ABSENT; one request returned no answer. HC achieves AUROC 0.777 (95\% CI $[0.70,0.84]$) and AUPRC 0.157, against 4.2\% event prevalence. We use stepwise average precision and 1,000 prompt-cluster bootstrap draws (seed 20260917) for AUROC uncertainty. Wan2.2 event rates are 0.0\% for Endpoint-free, 8.3\% for Full, and 5.0\% for Partial.

On 18 prompt--seed pairs across all nine conditions, native-frame judgments agree with judgments using 8 FPS input in 158/162 cases (97.5\%) and with those from a repeat evaluation in 159/162 (98.1\%).

\FloatBarrier
\subsection{Track Reversal Ratio and Back-Tracking Ratio}
\label{app:tracks}

\paragraph{Track extraction.}

CoTracker3 tracks points at the native 24 FPS on videos resized to $640\times352$ \citep{cotracker3}. We query a $24\times24$ grid at frames 0, 30, and 60, restricted to the region outside $\Mzero$ dilated by two latent cells. 

\paragraph{Reference-axis construction.}
Track $i$ is queried at native frame $t_i\in\{0,30,60\}$ of $T=121$ and has position $p_{i,t}$ at frame $t$. Its early reference velocity $v_i$ is the median of valid adjacent displacements during the first third of its remaining duration $T-t_i$; at least six valid steps are required. Tracks with $t_i<\lfloor T/2\rfloor=60$, that is, queried at frames 0 and 30, and $\|v_i\|\ge0.3$ pixels per frame provide the global reference
\begin{equation}
g=\frac{\sum_i\|v_i\|v_i}{\sum_i\|v_i\|},\qquad u_g=g/\|g\|.
\end{equation}
At least ten such tracks and $\|g\|\ge0.2$ pixels per frame are required. \paragraph{Track applicability.}
Each subsequently retained track must have at least four valid late steps and an along-axis excursion of at least six pixels; at least ten retained tracks are required.

\paragraph{Trajectory diagnostics.}
Writing $q_{i,t}=\langle p_{i,t}-p_{i,t_i},u_g\rangle$ for projected displacement and $\Delta q_{i,t}=q_{i,t+1}-q_{i,t}$ for a step between adjacent visible frames, TRR measures the fraction of along-axis step mass that is negative in the last third of the video, $t\ge\lfloor 2T/3\rfloor=80$:
\begin{equation}
\mathrm{TRR}=\frac{\sum_{i,t\in\mathcal L_i}\max(0,-\Delta q_{i,t})}{\sum_{i,t\in\mathcal L_i}|\Delta q_{i,t}|}.
\end{equation}
Here $\mathcal L_i$ includes only valid late adjacent steps. BTR measures return from the largest achieved displacement to the last visible track position $q_{i,\mathrm{end}}$, which need not be the video's last frame. Write $q_{i,\max}=\max_t q_{i,t}$ and $q_{i,\min}=\min_t q_{i,t}$ over visible positions. The per-track ratio and its excursion-weighted aggregation are
\begin{align}
e_i&=q_{i,\max}-q_{i,\min},\qquad
r_i^{\rm return}=\frac{q_{i,\max}-q_{i,\rm end}}{e_i},\\
\mathrm{BTR}&=\frac{\sum_i e_i r_i^{\rm return}}{\sum_i e_i}.
\end{align}
We add $10^{-6}$ to normalization denominators for numerical stability. TRR and BTR use point trajectories and a last-third late window, complementing BMR's flow-based last-half measurement.

\begin{table}[!tbp]
\centering
\setlength{\tabcolsep}{4pt}
\caption{Flow and track diagnostics on the common intersection of 398/515 Wan2.2-Turbo samples applicable under all three conditions. Every mean uses the same samples.}
\label{tab:tracks}
\fitbox{%
\begin{tabular}{lrrrr}
\toprule
Condition & Common applicable & BMR $\downarrow$ & TRR $\downarrow$ & BTR $\downarrow$ \\

\midrule
Endpoint-free & 398 / 515 & 0.386 & 0.385 & 0.332 \\

Full & 398 / 515 & 0.633 & 0.598 & 0.600 \\

Partial & 398 / 515 & 0.415 & 0.410 & 0.359 \\

\bottomrule
\end{tabular}}
\end{table}

\paragraph{Common-population comparison.}
Track diagnostics apply to 500/515 Endpoint-free, 413/515 Full, and 492/515 Partial videos. We compare all three metrics on their common 398 prompt--seed pairs across 98 prompts (Table~\ref{tab:tracks}). Partial--Full differences are $-0.218$, $-0.188$, and $-0.241$ in BMR, TRR, and BTR. Pooled Spearman correlations between BMR and TRR/BTR are 0.64/0.65 across the 1,194 paired, condition-specific observations. Main BMR results use all 515 videos per condition.

\subsection{BMR Measurement Region and Static Backgrounds}
\label{app:bmr_regions}

Main BMR is measured outside $\Mzero$, where Partial leaves the terminal frame unconstrained. Measuring the same quantity over the whole frame, or only inside $\Mzero$ where Full and Partial impose the same edited content, gives the same conclusions: Partial lowers BMR relative to Full on all three backbones, and donor context lowers it relative to Original Full on both tested backbones, in every region (Table~\ref{tab:bmr_regions}). Because the reference axis is estimated from early background motion, we also exclude prompt--seed pairs whose early reference motion $\|b_{\rm ref}\|$ is below one pixel per 8-FPS frame pair in either condition (25--39\% of pairs). Partial--Full then remains $-0.215$ $[-0.269, -0.162]$ on Wan2.2, $-0.278$ $[-0.331, -0.226]$ on SANA, and $-0.298$ $[-0.355, -0.243]$ on Hunyuan, and donor context minus Original Full remains $-0.251$ $[-0.303, -0.201]$ on Wan2.2 and $-0.232$ $[-0.287, -0.176]$ on Hunyuan.

\begin{table}[!tbp]
\centering
\setlength{\tabcolsep}{4pt}
\caption{BMR differences by measurement region (103 prompts $\times$ five seeds). Outside $M_0$ is the main BMR; the whole-frame and inside-$M_0$ variants apply Eq.~\eqref{eq:bmr}, including its reference axis, to all pixels or to pixels inside $M_0$.}
\label{tab:bmr_regions}
\fitbox{%
\begin{tabular}{@{}lccc@{}}
\toprule
Comparison & Outside $M_0$ (main) & Whole frame & Inside $M_0$ \\
\midrule
Wan2.2-Turbo, Partial $-$ Full & $-0.213^{*}\;[-0.262, -0.165]$ & $-0.202^{*}\;[-0.251, -0.154]$ & $-0.141^{*}\;[-0.187, -0.097]$ \\
SANA-Video, Partial $-$ Full & $-0.285^{*}\;[-0.325, -0.246]$ & $-0.289^{*}\;[-0.329, -0.250]$ & $-0.281^{*}\;[-0.324, -0.238]$ \\
HunyuanVideo-1.5, Partial $-$ Full & $-0.252^{*}\;[-0.296, -0.208]$ & $-0.248^{*}\;[-0.292, -0.205]$ & $-0.223^{*}\;[-0.272, -0.176]$ \\
Wan2.2-Turbo, D $-$ O & $-0.238^{*}\;[-0.283, -0.195]$ & $-0.230^{*}\;[-0.275, -0.186]$ & $-0.143^{*}\;[-0.188, -0.099]$ \\
HunyuanVideo-1.5, D $-$ O & $-0.227^{*}\;[-0.267, -0.186]$ & $-0.218^{*}\;[-0.259, -0.178]$ & $-0.183^{*}\;[-0.225, -0.142]$ \\
\bottomrule
\end{tabular}}
\end{table}

\begin{table}[!tbp]
\centering
\setlength{\tabcolsep}{4pt}
\caption{Outside-$\Mzero$ first--last-frame RGB difference of preserved context: mean absolute RGB difference between each video's first and last frames. For reference, FLUX.2 endpoints differ from the source image by 0.056 $[0.045, 0.068]$ outside $\Mzero$ and 0.232 $[0.206, 0.258]$ inside. Intervals use the paired prompt bootstrap. Wan2.2-Turbo Partial videos were regenerated with identical settings for this measurement.}
\label{tab:layout}
\fitbox{%
\begin{tabular}{@{}lccccc@{}}
\toprule
 & \multicolumn{3}{c}{First--last RGB difference} & \multicolumn{2}{c}{Difference from Endpoint-free} \\
\cmidrule(lr){2-4} \cmidrule(l){5-6}
Model & Endpoint-free & Full & Partial & Full & Partial \\
\midrule
Wan2.2-Turbo & 0.177 & 0.076 & 0.164 & $-0.102\;[-0.116, -0.087]$ & $-0.014\;[-0.019, -0.009]$ \\
SANA-Video & 0.233 & 0.058 & 0.216 & $-0.174\;[-0.195, -0.155]$ & $-0.017\;[-0.024, -0.010]$ \\
HunyuanVideo-1.5 & 0.127 & 0.082 & 0.126 & $-0.045\;[-0.059, -0.031]$ & $-0.001\;[-0.002, +0.000]$ \\
\bottomrule
\end{tabular}}
\end{table}

\subsection{Source-Time Layout of Preserved Context}
\label{app:layout}
We measure the first--last-frame RGB difference of preserved context outside the estimated support. Images and frames are downsampled to width 640, and $\Mzero$ is resized using nearest-neighbor interpolation, as for BMR. The primary distance is the mean absolute RGB difference in $[0,1]$ outside $\Mzero$; LPIPS-Alex is secondary \citep{lpips}. For videos, we compare each video's first and last frames at its own resolution, avoiding resampling differences between pipelines. In 93\% of prompts, FLUX.2 endpoints differ from the source image less outside $\Mzero$ than inside. Outside $\Mzero$, in 91--99\% of prompts, Endpoint-free videos change more from their first to their last frame than the endpoint differs from the source. Under Full, the last frame stays close to the first frame outside $\Mzero$, and Partial leaves the difference within 0.02 of Endpoint-free (Table~\ref{tab:layout}). LPIPS gives the same ordering.

Before computing these quantities, we also specified a dose--response test. Across prompts, we correlate the Endpoint-free first--last-frame difference with Full's HC and BMR increases over Endpoint-free and with Partial's reductions relative to Full (Spearman correlation with a prompt bootstrap). For Full$-$Endpoint-free, the difference and the outcomes come from disjoint seeds. Only the two Hunyuan HC tests show a dose--response relation ($\rho=+0.28\;[+0.08,+0.47]$ and $-0.35\;[-0.50,-0.17]$); the other ten intervals include zero. We therefore do not claim that Full's artifacts scale with scene motion.

\clearpage
\section{Residual Failures and Exploratory Support Release}
\label{app:failures}

\begin{figure}[htbp]
\centering
\includegraphics[width=\linewidth]{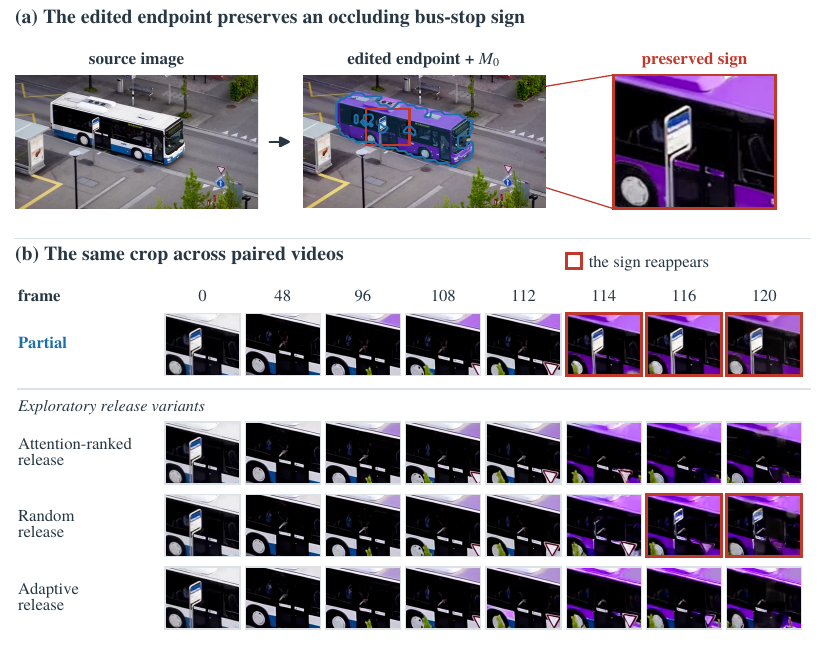}
\caption{\textbf{A residual failure of fixed support.} (a) The edited endpoint preserves an occluding bus-stop sign; blue shows $M_0$, and the red box defines the crop. (b) Identical image-coordinate crops at all eight original frame indices. Red borders mark the sign's reappearance in Partial and random release; attention-ranked and adaptive release avoid it in this example. The release variants are exploratory. All rows use Wan2.2-Turbo, purple-bus case 90001 and seed 22001, as in Fig.~\ref{fig:fixed_failure}; frame intervals are unequal.}
\label{fig:residual}
\end{figure}

Fixed support cannot follow moving content. Figure~\ref{fig:residual} shows an endpoint-preserved sign reappearing under Partial after disappearing earlier in the video.

\paragraph{Release rules.}
We explore shrinking support over Wan2.2's denoising timesteps $1000,960,889,727$. Adaptive release starts from the pooled, fractional $M_0$ weights $w_0$ on the $22\times40$ token grid. At step $k$, we compute the channelwise Euclidean distance between the endpoint latent and the clean prediction of every latent slot except the first and last, average across these slots, and pool spatially by $2\times2$ to obtain $G_k$. With $\tau_k$ the mean of $G_k$ over all tokens, we set $w_{k+1}=w_k\mathbf1[G_k<\tau_k]$ after each of the first three steps; released tokens never re-enter.

Random and attention-ranked release rank the tokens inside $M_0$ randomly or by pooled FLUX.2 attention. At each step, they keep tokens in rank order, with their pooled weights, while the cumulative weight stays within the adaptive run's total for that sample and step; tokens outside $M_0$ are never constrained. The same $w_k$ controls model-input blending, token timesteps, and clean-prediction projection.

\paragraph{Completion--discontinuity trade-off.}
On the 515-video Wan2.2 cohort, adaptive release yields TCR 38.45\%, TC-Score 0.811, HC 4.89, and DD 76.3\%, compared with fixed support's 42.91\%, 0.819, 5.76, and 76.7\%. Attention-ranked and adaptive release avoid the illustrated sign reappearance, while random release does not. The aggregate adaptive result trades lower HC for lower TCR; this example does not establish a general advantage over fixed support.

\FloatBarrier
\section{Broader I2V Completion Comparisons}
\label{app:overview}

\begin{table*}[!htbp]
\centering
\caption{Automatic evaluation results of I2V models on TC-Bench-I2V. FPS reports video-stage throughput.}
\label{tab:tcbench_i2v}
\setlength{\tabcolsep}{4pt}
\renewcommand{\arraystretch}{1.15}
\renewcommand{\venue}[1]{{\scriptsize\textcolor{black!55}{#1}}}
\fitbox{%
\begin{tabular}{@{}lcccccccccc@{}}
\toprule
 & & \multicolumn{8}{c}{\textbf{TC-Bench-I2V} $\uparrow$} & \\
\cmidrule(lr){3-10}
\textbf{Methods} & \textbf{Params} & \multicolumn{2}{c}{\textbf{Attribute}} & \multicolumn{2}{c}{\textbf{Object}} & \multicolumn{2}{c}{\textbf{Background}} & \multicolumn{2}{c}{\textbf{Overall}} & \textbf{FPS} $\uparrow$ \\
\cmidrule(lr){3-4}\cmidrule(lr){5-6}\cmidrule(lr){7-8}\cmidrule(lr){9-10}
 & & \textbf{TCR} & \textbf{TC-Score} & \textbf{TCR} & \textbf{TC-Score} & \textbf{TCR} & \textbf{TC-Score} & \textbf{TCR} & \textbf{TC-Score} & \\
\midrule
\multicolumn{11}{@{}l}{\textit{\textbf{Image-to-Video Generation}}} \\
Wan2.2-Turbo \venue{Release'25.08} & 5B & 31.60 & 0.7647 & \underline{34.74} & 0.7842 & 38.67 & 0.8486 & 33.79 & 0.7841 & 2.270 \\

Wan2.2 Base \citep{wan} \venue{Release'25.07} & 5B & 32.00 & 0.7935 & 30.00 & 0.7794 & 30.67 & 0.8442 & 31.07 & 0.7957 & 0.182 \\

SANA-Video \citep{sana} \venue{ICLR'26} & 2B & 23.20 & 0.7729 & 22.11 & 0.7245 & 28.00 & 0.7590 & 23.50 & 0.7530 & 1.031 \\

HunyuanVideo-1.5 \citep{hunyuan} \venue{arXiv'25.11} {\scriptsize\textcolor{black!55}{(480p, 8-step)}} & 8.3B & 52.80 & \underline{0.9076} & 25.26 & \textbf{0.8312} & 41.33 & 0.8556 & 40.97 & \underline{0.8718} & 0.615 \\

MfM \citep{mfm} \venue{ICLR'26} & 8.1B & 39.60 & 0.8449 & 24.74 & 0.7887 & 29.33 & 0.8365 & 32.62 & 0.8230 & 0.034 \\

OmniWeaving \citep{omniweaving} \venue{arXiv'26.03} & 8.3B & 42.80 & 0.8653 & \textbf{35.79} & 0.8034 & 37.33 & 0.8305 & 39.42 & 0.8374 & 0.027 \\

\midrule
\multicolumn{11}{@{}l}{\textit{\textbf{Image-to-Video Generation w/ Image Editing (Full)}}} \\
FLUX.2 + Wan2.2-Turbo & 14B & 48.40 & 0.8466 & 31.58 & 0.8081 & 46.67 & 0.8655 & 41.94 & 0.8352 & 2.270 \\

ChronoEdit + Wan2.2-Turbo & 19B & 40.80 & 0.8239 & 33.68 & 0.7891 & \textbf{54.67} & \underline{0.8948} & 40.19 & 0.8214 & 2.266 \\

Qwen-Image-Edit + Wan2.2-Turbo & 25B & 41.20 & 0.8096 & 30.53 & 0.7736 & 26.67 & 0.8357 & 35.15 & 0.8001 & 2.266 \\

FLUX.2 + Wan2.2 Base & 14B & 54.80 & 0.8700 & 31.58 & 0.8034 & \underline{48.00} & 0.8838 & \underline{45.24} & 0.8475 & 0.186 \\

FLUX.2 + SANA-Video & 11B & 44.40 & 0.8451 & 25.79 & 0.7200 & 32.00 & 0.8136 & 35.73 & 0.7944 & 1.029 \\

FLUX.2 + HunyuanVideo-1.5 & 17.3B & \textbf{58.40} & 0.9018 & 31.05 & 0.8069 & 46.67 & 0.8470 & \textbf{46.60} & 0.8588 & 0.565 \\

FLUX.2 + MfM & 17.1B & \underline{58.00} & \textbf{0.9344} & 25.79 & 0.8198 & 42.67 & \textbf{0.8983} & 43.88 & \textbf{0.8869} & 0.034 \\

FLUX.2 + OmniWeaving & 17.3B & 51.60 & 0.8977 & 32.63 & \underline{0.8200} & 42.67 & 0.8683 & 43.30 & 0.8647 & 0.029 \\

\bottomrule
\end{tabular}}
\end{table*}

We additionally compare ChronoEdit and Qwen-Image-Edit endpoints \citep{chronoedit,qwenimage} on the same 103 prompts and five seeds. Image-editing rows use Full conditioning. Tables~\ref{tab:subgroups} and~\ref{tab:endpoint_tcr} provide paired completion differences; Table~\ref{tab:main} evaluates Partial.

\paragraph{Endpoint sources and parameter counts.}
The Params column reports generative-backbone parameter counts in billions, including the editor but excluding separate text/vision encoders and VAEs. FLUX.2 uses klein-base-9B and the final-frame prompt in Appendix~\ref{app:implementation}. ChronoEdit uses its 14B checkpoint, which formulates editing as video generation with the input and edited images as the first and last frames. We therefore supply the original transition prompt directly, without the final-frame template, and use the last of five generated frames as the endpoint. Thus both editor and prompt formulation change. Qwen-Image-Edit uses the 20B 2511 release\footnote{Official model card: \href{https://huggingface.co/Qwen/Qwen-Image-Edit-2511}{Qwen-Image-Edit-2511}.} with the FLUX.2 final-frame prompt, so only the editor changes.

\paragraph{Native configurations and timing.}
MfM uses 97 frames at $1280\times704$ and OmniWeaving 81 at 720p. FPS is computed as the number of decoded frames divided by the median video-stage time, excluding endpoint generation. MfM includes batch-amortized startup; other methods use per-video timers.

\end{document}